\documentclass{article}

\usepackage[nonatbib,preprint]{neurips_2026}
\usepackage[square,numbers]{natbib}
\usepackage[utf8]{inputenc} 
\usepackage[T1]{fontenc}    
\usepackage{hyperref}       
\usepackage{url}            
\usepackage{amsfonts}       
\usepackage{nicefrac}       
\usepackage{microtype}      
\usepackage{xcolor}         

\usepackage{graphicx}
\usepackage{subcaption}
\usepackage{booktabs} 
\usepackage{enumitem}
\usepackage{multirow}
\usepackage{longtable}
\usepackage{threeparttable} 
\usepackage{makecell}
\usepackage{placeins} 
\usepackage{wrapfig}

\usepackage{amsmath}
\usepackage{amssymb}
\usepackage{mathtools}
\usepackage{amsthm}

\usepackage[capitalize,noabbrev]{cleveref}

\theoremstyle{plain}

\theoremstyle{definition}

\theoremstyle{remark}

\usepackage[textsize=tiny]{todonotes}

\title{GRINQH: Graded Input-based Quantization Hierarchy for Efficient LLM Generation}

\author{
  Jette Oberländer~$^{1,3}$\thanks{These authors contributed equally to this work.}
  \hfill
  Jan Finkbeiner~$^{2,3}$\footnotemark[1]
  \hfill
  Catherine M.~Schöfmann~$^{2,3}$
  \hfill
  Emre Neftci~$^{2,3}$
  \\[2.0em]
    $^1$ Fakultät für Informatik, RWTH Aachen, Aachen, 52074, Germany \\
    $^2$ Fakultät für Elektrotechnik und Informationstechnik, RWTH Aachen, Aachen, 52062, Germany\\
    $^3$ Peter Grünberg Institut, Forschungszentrum Jülich GmbH, Jülich, 52425, Germany\\
  \{j.oberlaender, j.finkbeiner, c.schoefmann, e.neftci\}@fz-juelich.de 
}

\begin{document}

\maketitle

\begin{abstract}
Autoregressive decoding with LLMs is primarily bottlenecked by GPU memory bandwidth,
especially in edge-computing settings. 
While quantization is essential for mitigating this bottleneck, most existing methods treat inference as a uniform process and fail to account for the asymmetry between the compute-bound prefill stage and the memory-bound decoding stage.
We propose GRINQH (GRaded INput-based Quantization Hierarchy), a weight-only post-training quantization framework that accelerates decoding by unifying quantization and sparsification. 
GRINQH leverages activation magnitudes as a proxy for computational importance to dynamically assign weight channels to different precision levels, enabling flexible average bit widths during decoding. 
Evaluated on Llama3 and Qwen3 models, GRINQH outperforms state-of-the-art fixed- and mixed-precision baselines at comparable 3- and 4-bit settings, even enabling effective 2-bit generation. 
We experimentally verify theoretical speedups by leveraging a hierarchical nested memory layout for multi-precision storage in a custom GPU kernel. 
Ultimately, GRINQH establishes a new state-of-the-art Pareto frontier for LLM generation, enabling a dynamic trade-off between generation quality and inference speed. 
\end{abstract}

\begin{wrapfigure}[15]{r}{0.52\linewidth}
    \vspace{-1.3em}
    \centering    
    \includegraphics[width=1.\linewidth]{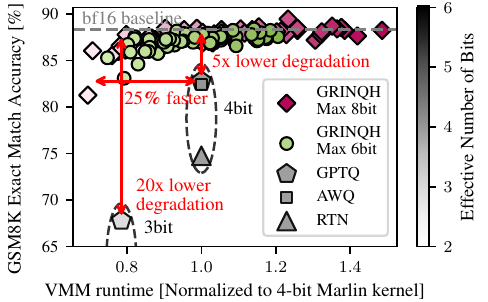}
    \caption{
      Pareto frontier of GSM8K accuracy vs. VMM runtime for Qwen3-8B on an RTX 4090. GPTQ, AWQ, RTN assume MARLIN execution.
    }
    \label{fig:gsm8k_over_RTX4090}
\end{wrapfigure}
\section{Introduction}
The primary operational focus for Large Language Models (LLMs) has shifted from training to efficient inference at scale. Despite advances in hardware throughput, inference remains constrained by the memory wall: a fundamental performance bottleneck where off-chip DRAM transfer speeds cannot keep pace with on-chip computation. In LLMs, this is driven by dense linear layers, which, for models exceeding 7B parameters, account for approximately 95\% of total parameters and 65--85\% of total floating-point operations (FLOPs) \cite{Dettmers22}. 
We propose GRINQH, a weight-only quantization framework that directly targets these linear layers to accelerate inference by reducing the volume of weight data transferred from memory. By enabling fine-grained, channel-wise (row-by-row) dynamic precision adjustment, GRINQH facilitates structured sparse memory loads that significantly reduce weight data movement.
This approach is motivated by the fact that LLM inference operates in two fundamentally different stages: prefill and decoding. 
In the prefill stage, the input prompt is processed in parallel, allowing model weights to be loaded once and reused across the batch, rendering this stage largely compute-bound \cite{Pope22}. 
However, the bottleneck shifts to a memory-bound stage during autoregressive decoding, where all weight matrices must be loaded from DRAM to generate a single token. 
Since these matrices comprise the vast majority of the model’s footprint, loading them repeatedly for each token is highly inefficient.
Especially for edge applications with batch size one, GPU compute units remain underutilized while waiting for data from DRAM, resulting in latencies significantly higher than in the prefill stage \cite{Pope22, Lin23, Liu23}. 

Various compression techniques have been proposed to mitigate this memory-bound bottleneck, including quantization \cite{Lin23, Frantar22, Xiao22}, pruning \cite{Sun23}, and sparsification \cite{Mirzadeh23, Song24, Liu24}. 
Activation sparsity, in particular,
dynamically bypasses weight loading based on runtime activations \cite{Liu23, Song24}. 
However, these methods often require costly retraining or fail to sufficiently address the memory wall \cite{Mirzadeh23, Liu24}.
In contrast, post-training quantization (PTQ) \cite{Lin23, Frantar22, Xiao22} is a highly effective strategy for accelerating inference while maintaining competitive model accuracy without significant overhead.
Standard quantization methods, however, typically apply a uniform strategy across both stages. 
Low precision during the compute-bound prefill stage provides limited speedup, as performance is bottlenecked by arithmetic throughput rather than memory transfer, yet it introduces quantization noise that unnecessarily degrades accuracy. 
This mismatch between algorithmic design and hardware behavior leaves substantial efficiency gains untapped.

In this work, we argue that efficient LLM inference requires treating decoding as the primary optimization target. 
Our decoding-centric framework, GRINQH, realizes this by explicitly decoupling the prefill and decoding stages. 
Specifically, GRINQH preserves high precision during prefill to maintain model accuracy while dynamically adjusting weight precision (0--8 bits) during decoding based on activation magnitudes.
A custom bit-planar memory layout ensures that only the required bits are fetched from DRAM for each row, directly alleviating the bandwidth bottleneck.
Our custom GPU decoding kernel shows that latency scales smoothly with effective bits and achieves performance competitive with optimized fixed-precision kernels such as MARLIN \cite{frantar24marlin}. 
Furthermore, assigning higher precision to weights associated with high-magnitude activations (outliers) \cite{Lin23}, while aggressively reducing precision elsewhere, leads to significantly improved accuracy compared to quantizing to a single precision level. 
The inherent nested weight structure also provides GRINQH with elastic multi-precision capability, enabling real-time trade-offs between generation quality and speed within a single deployed model.

Through synergies between our algorithmic framework and efficient GPU kernels, GRINQH not only outperforms existing methods at iso-bit precision but also defines a new Pareto frontier for the trade-off between generation quality and generation speed (Fig.~\ref{fig:gsm8k_over_RTX4090}).
Our key contributions are as follows:
\begin{itemize}[nosep]
    \item \textbf{Dynamic PTQ Framework:} We introduce GRINQH, a fine-grained, dynamic post-training quantization (PTQ) framework that targets the memory-bound decoding stage of LLM inference by loading weight channels in varying precisions depending on the magnitude of real-time input activations.
    \item  \textbf{Superior Accuracy-Efficiency Trade-offs:} We show that GRINQH consistently outperforms state-of-the-art static PTQ methods, such as AWQ and GPTQ, in terms of accuracy at comparable effective bit widths, particularly for 2-, 3- and 4-bit quantization. In doing so, our method establishes a new Pareto frontier in generation quality versus generation speed.
    \item  \textbf{Optimized GPU Kernels:} We develop a custom GPU decoding kernel that scales gracefully with the number of effective bits, achieving performance competitive with optimized fixed-precision quantization kernels.
\end{itemize}
\section{Related Work}

\textbf{Quantization and Sparsification.}
Weight quantization and activation sparsification are two widely explored techniques for reducing memory traffic in LLMs. 
Fundamentally, sparsification can be viewed as an extreme case of quantization where the allocated bit width is reduced to zero (0-bit). 
Efficient hardware acceleration in this regime requires structured sparsity, such as zeroing out entire weight channels or blocks, to skip memory loads effectively.
While activation sparsity naturally induces these patterns by allowing the hardware to bypass weight channels corresponding to zero-valued activations, training-free sparsity induction methods like magnitude-based thresholding \cite{Liu24} often yield insufficient memory savings or significant accuracy degradation. 
Consequently, sparsification alone is often inadequate to mitigate the memory bottleneck in modern, dense LLMs.

\textbf{Post-Training Quantization (PTQ).}
Unlike sparsification, quantization reduces memory traffic by representing weights in reduced bit widths rather than skipping them entirely. 
PTQ is particularly effective as it avoids costly training, but its accuracy is fundamentally challenged by high-magnitude outliers in LLM activations and weights. 
Early mixed-precision approaches attempt to protect critical weight outliers by keeping salient weights in high precision \cite{Dettmers22}, yet they often struggle to translate theoretical savings into real-world speedups. 
Subsequent research demonstrated that managing activation outliers is often more crucial for maintaining model performance \cite{Lin23}. 
Consequently, fixed-bit-width methods like GPTQ \cite{Frantar22} and AWQ \cite{Lin23} emerged to protect outliers through second-order information or scaling, while techniques such as QuaRot \cite{Ashkboos23} and QuIP \cite{Tseng24quip} suppress outlier effects via basis transformations. 
However, these methods overlook the divergent hardware bottlenecks of prefill and decoding, and are constrained by applying a single, static precision level regardless of weight importance. 
To address this, recent mixed-precision research explores three primary paradigms: (i) nested weight representations \cite{AnyPrecisionLLM24,Nair25,MatGPTQ26}, (ii) static allocation at the layer or block level \cite{MatFormer23, Nair25, MatGPTQ26,SliM-LLM24}, and (iii) coarse-grained dynamic scheduling, such as per-token precision adjustment \cite{Chen24PMPD}.
While nested frameworks like any-precision LLM \cite{AnyPrecisionLLM24} offer theoretical flexibility, they often lack the low-level loading logic required for mixed-precision execution.
Similarly, static methods \cite{SliM-LLM24} cannot adapt to the runtime emergence of activation outliers, and existing dynamic approaches operate only on a coarse-grained level, e.g., at the token level \cite{Chen24PMPD}, failing to isolate and protect fine-grained, channel-wise activation outliers.
To address these limitations, GRINQH introduces a hardware-verified framework that pairs the efficiency of nested weights with stage-aware, row-wise precision adjustment.

 \textbf{Positioning of GRINQH.}
GRINQH unifies sparsification and quantization within a decoding-centric, input-adaptive framework. 
Unlike prior static allocation strategies, GRINQH dynamically fetches mixed-precision weights at each decoding step based on real-time activation magnitudes. 
By assigning zero bits to negligible channels and higher precision to salient ones, GRINQH protects outliers while maximizing effective sparsity. 
By decoupling the memory-bound decoding stage from prefill, our framework resolves the algorithm-hardware mismatch prevalent in current literature. 
GRINQH serves as a flexible execution layer compatible with various optimization and quantization methods (e.g., GPTQ), translating theoretical bit-savings into measurable hardware acceleration.
\begin{figure*}[t]
  \begin{center}
    \centerline{\includegraphics[width=\textwidth]{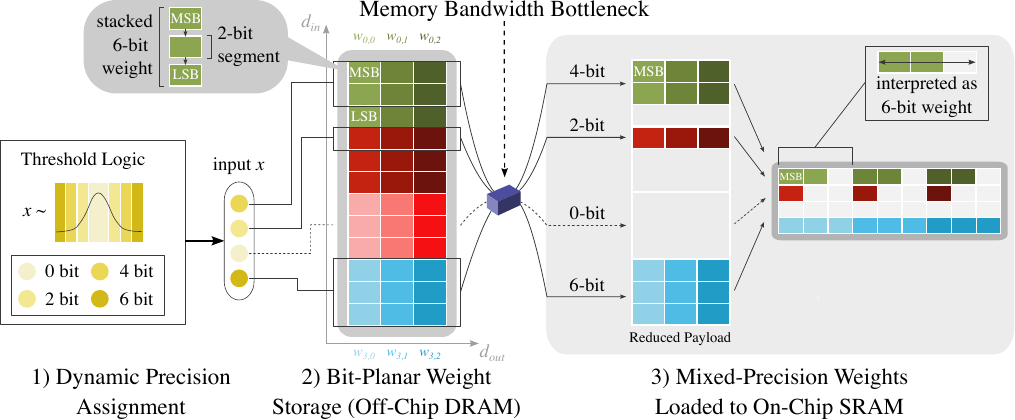}}
    \caption{
      \textbf{Overview of GRINQH.} During decoding, GRINQH mitigates the memory bandwidth bottleneck through a dynamic, channel-wise precision loading scheme. \textbf{1) Precision Assignment:} Input activations $x_i$ are mapped to bit widths $b_i \in \{0,2,4,6\}$ based on their magnitude via pre-computed thresholds from a calibration set.
      \textbf{2) Bit-Stacked Storage:}  Weights are stored in DRAM using a bit-planar format. Each $b_\text{max}=6$-bit weight is decomposed into $b_\text{max} / 2$ 2-bit planes which are stacked along the input-channel dimension ($d_\text{in}$). Weights of the same input channel are contiguously aligned along the output-channel axis ($d_\text{out}$), ensuring efficient, coalesced loading of individual input channels. \textbf{3) On-Chip Reconstruction:} Based on the assigned precision $b_i$ for each weight channel, only the most significant segments are fetched through the bandwidth-constrained DRAM interface. These segments are reconstructed on-chip into a mixed-precision weight matrix, 
      significantly reducing DRAM-to-SRAM traffic while maintaining high model performance.
    }
    \label{fig:GRINQH}
  \end{center}
\end{figure*}
\section{GRINQH: Graded Input-based Quantization Hierarchy}  
We propose GRINQH, a weight-only post-training quantization (PTQ) framework designed to mitigate the memory-bandwidth bottleneck in sequential LLM decoding. 
GRINQH is built upon two synergistic pillars: (i)~an input-dependent precision assignment that adaptively allocates bit widths to weight channels and (ii)~a hierarchical bit-planar memory layout that enables low-overhead extraction of multi-precision weights (Fig.~\ref{fig:GRINQH}).

To optimize end-to-end throughput, GRINQH differentiates between execution stages based on their primary hardware constraints.
During the prefill stage, which is typically compute-bound, weights are loaded at a static, fixed precision.
This design choice maintains the high arithmetic intensity required for high-throughput prompt processing. 
Conversely, during autoregressive decoding with a batch size of one, the system becomes primarily memory-bandwidth bottlenecked (see roofline analysis in Appendix~\ref{app:rooflineAnalysis}). 
In this stage, latency is dominated by the time required to transfer weight data from off-chip DRAM to on-chip SRAM. 
This ``memory wall'' arises as LLM weights far exceed the capacity of on-chip SRAM, necessitating a full transfer of the entire weight matrix for every generated token.
To mitigate this constraint, GRINQH dynamically evaluates the magnitude of incoming activations at each linear layer to determine the optimal precision for the corresponding weight channels. 
By loading weight channels linked to high-magnitude ``outlier'' activations at high precision while reducing the precision of, or entirely bypassing the loading of channels with low-magnitude signals, GRINQH significantly reduces DRAM-to-SRAM traffic while preserving accuracy. 
This precision redistribution prioritizes the model’s most influential representational components, recovering quantization accuracy while providing fine-grained, per-channel control over total memory throughput.
\subsection{Dynamic Precision Assignment} 
During the decoding stage, we assign a bit width $b_i \in \{b_0, b_1, \dots, b_n\}$ ($b_n = b_\text{max}$ represents the maximum available precision) to each activation element $x_i$ along the input dimension ($i\in \{0,\ldots,d_\text{in}-1\}$) of the current token. This assignment is determined by the magnitude $|x_i|$ relative to calibrated thresholds, which dictates the precision of the corresponding weight channel $\mathbf{w}_{i,:}$ (see color-coded precision distribution in Fig.~\ref{fig:GRINQH}).

Motivated by LLM sensitivity to activation outliers~\cite{Xiao22} and the existence of ``free sparsity''~\cite{Liu24}, 
GRINQH protects influential activations with high-precision weights while processing background channels with lower precision or bypassing them entirely.

Formally, for each layer $\ell$, the bit width $b_i$ is assigned as $b_i = b_k$ if $|x_i| \in [\theta_{k}^{(\ell)}, \theta_{k+1}^{(\ell)})$, where $k \in \{0, \dots, n\}$ and the boundary thresholds are defined as $\theta_0^{(\ell)} = 0$ and  $\theta_{n+1}^{(\ell)} = \infty$.
By allocating high precision only to channels associated with large activations, GRINQH effectively implements dynamic activation outlier protection. 
Unlike static quantization or smoothing methods limited by fixed calibration statistics, GRINQH adapts to shifting outliers at each time step. 
The dynamic precision allocation induces dynamic structured sparsity that reduces memory traffic and accelerates inference without the accuracy trade-offs inherent to static pruning or quantization.

\subsection{Threshold Calibration} 
To minimize runtime overhead, the thresholds $\Theta$ are derived offline via percentile mapping on a small, representative calibration dataset (see Appendix~\ref{app:RepresentativenessCalibration} for an analysis of data robustness and resource consumption). 
We define a global target precision distribution $\mathrm{P} = \left(p_0,p_1,\dots, p_n\right)$, where each $p_k$ represents the desired proportion of weight channels to be processed at bit width $b_k$.
For each layer $\ell$, we compute the empirical percentiles that satisfy $\mathrm{P}$ for every sample in the calibration set. These values are averaged to produce a single set of static thresholds $\Theta^{(\ell)}$, where $\theta_k^{(\ell)}$ corresponds to the ($\sum_{j=0}^{k-1} p_j$)-th percentile of the empirical activation magnitudes. 

By pre-calculating these thresholds, the runtime ``grading'' logic is reduced to a set of static scalar comparisons against the incoming activation vector. This ensures that dynamic precision assignment adds negligible latency to the inference pipeline, allowing our multi-precision extraction to maintain performance parity with highly optimized static kernels (see Fig.~\ref{fig:kernel_gemv_bench}).
Further discussion regarding the selection and impact of different precision distributions $\mathrm{P}$ is provided in Appendix~\ref{app:HyperparameterSelection}.
\subsection{Hierarchical Bit-Slicing} \label{sec:hierarchicalBitSlicing}
GRINQH operates as a quantization-agnostic framework, optimizing bit-width distribution to surpass the baseline performance of the underlying quantization primitive. To realize dynamic multi-precision loading without the memory overhead of storing multiple weight copies, we employ a \textit{hierarchical bit-planar interleaved storage} format. 

\textbf{Storage Layout and Weight Loading.}
This format requires the underlying quantization to satisfy a nested property, which is naturally met by uniform methods and specific non-uniform techniques such as k-means quantization~\cite{AnyPrecisionLLM24}.
Under uniform quantization, each weight $w_{ij}$ is quantized to a $b_\text{max}$-bit integer $q_{b_\text{max}}$; for non-uniform or inherently nested weights, the parent model must match $b_\text{max}$ to ensure compatibility. The quantized weight $q_{b_\text{max}}$ is decomposed into discrete 2-bit segments representing successive refinements of the value (see \textit{Bit-Planar Weight Storage} in Fig.~\ref{fig:GRINQH}). This organization enables the kernel to fetch weights in 2-bit increments, supporting an adaptive range of $b \in \{0,2,\dots, b_\text{max}\}$.

While Fig.~\ref{fig:GRINQH} illustrates a configuration with $b_\text{max}=6$, the framework is indifferent to the choice of maximum bit width; we evaluate $b_\text{max} \in \{4,6,8\}$ to balance precision range against storage efficiency.
To facilitate high-speed vector-matrix multiplication, weights are stored in a transposed, input-major layout. This ensures that for any element (or block) of the input vector $\mathbf{x}$, the corresponding weight channels for all output dimensions $d_\text{out}$ are row-contiguous. 
To access a target precision $b \leq b_\text{max}$, the kernel only loads the $\nicefrac{b}{2}$  most significant segments from DRAM to on-chip SRAM, effectively reducing memory traffic while providing the bits required for reconstruction (see Appendix~\ref{app:kernelImplementationDetails} for more details).

\textbf{Weight Reconstruction and Bias Correction.}
For uniform quantization, let $s$ denote the quantization scale factor and $z$ the zero-point  correction determined for the full $b_\text{max}$ precision. The dequantized floating-point value $\hat{w}$ is reconstructed from the truncated $b$-bit integer $q_b$ as:
\begin{equation*}
\hat{w}= s \cdot ( q_b \cdot 2^{b_\text{max}-b} + \phi) + z ,
\end{equation*}
where $q_b$ is the integer formed by the concatenated segments and $\phi=2^{b_\text{max}-b-1}$ 
is a midpoint bias-correction term. Since bit-planar truncation acts as a floor operation, $\phi$ is essential to shift the reconstruction from the lower bound of the quantization bin to its center.
Our ablation studies confirm that without this correction, cumulative quantization bias across deep layers leads to representational collapse. 
Uncorrected models in our testing degraded to $0\%$ accuracy on GSM8K and exhibited WikiText-2 perplexity explosions exceeding $10^8$. Notably, for non-uniform nested weights utilizing Look-Up Tables (LUTs), this correction is unnecessary as the nested levels are inherently centered.

\section{Experimental Setup}
\label{sec:experimentalsetup}
We evaluate GRINQH across the Llama3 (1B, 3B, and 8B Instruct)~\cite{Grattafiori24} and Qwen3 (0.6B, 1.7B, 4B, and 8B)~\cite{Yang25Qwen3} model families to demonstrate its robustness across scales and architectures.
All accuracy-only evaluations are completed with the reference implementation of GRINQH, a PyTorch-level simulation of the kernel logic that allows for larger-scale batching. End-to-end results and all reported speedups are obtained using the Triton kernel, which is functionally equivalent to the reference but utilizes an alternative prefill path.

\textbf{Models and Baselines.} 
We compare GRINQH against several uniform quantization methods including Round-to-Nearest (RTN), AWQ~\cite{Lin23}, GPTQ~\cite{Frantar22}, QuaRot with GPTQ~\cite{Ashkboos24}, and AutoRound~\cite{SignRoundV123}, targeting 3-bit and 4-bit symmetric configurations. To reflect modern hardware trends, we also include the 4-bit NVFP4 format~\cite{abecassis_nvfp42025}. For mixed-precision and non-uniform comparisons, we evaluate against SliM-LLM~\cite{SliM-LLM24} and any-precision LLM~\cite{AnyPrecisionLLM24}. All methods employ a group size $G=128$, except NVFP4 ($G=16$) and any-precision LLM (per-channel). Further comparisons are detailed in Appendix~\ref{app:comparisonAgainstMultiprecisionMethods}.
GRINQH utilizes weights pre-quantized via RTN or GPTQ at $b_\text{max} \in \{4,6,8\}$ bits, as well as any-precision LLM at $b_\text{max}=8$. 
For all methods requiring calibration, we use 128 samples from The Pile (Uncopyrighted) dataset~\cite{Gao21thePile} with a 2048-token context window.
We conducted a hyperparameter sweep over the target precision distributions $\mathrm{P}$ (see Appendix~\ref{app:HyperparameterSelection} for full selection). 
Notably, we constrain the 0-bit (sparsity) proportion $p_0$ to a maximum of 30\%. 
Our empirical testing indicated that exceeding this threshold often led to significant performance degradation.

\textbf{Implementation and Deployment.} 
We perform RTN, AWQ, GPTQ, QuaRot, AutoRound, and NVFP4 quantization using the llm-compressor library~\cite{llmcompressor24}. For SliM-LLM, any-precision LLM, and PMPD~\cite{Chen24PMPD}, we use the official repositories.
Baselines are deployed via \texttt{vLLM}~\cite{kwon23vllm} for 4-bit models and HuggingFace Transformers ~\cite{Wolf19HuggingfaceTransformers} for 3-bit configurations, as \texttt{vLLM} currently lacks native 3-bit kernel support. 
To deploy GRINQH, we extend the \texttt{vLLM} backend with custom functionality for dynamic precision assignment, selective bit-planar memory access, and weight reconstruction.

\textbf{Evaluation Benchmarks.} 
We evaluate all models for accuracy using the EleutherAI LM Evaluation Harness\cite{Sutawika25LMeval} across seven benchmarks: WikiText-2 (Wiki2), LAMBADA (LMB), MMLU, BoolQ, HellaSwag, ARC-Challenge, and GSM8K (Chain-of-Thought, CoT) \cite{Merity16wikitext, Paperno16lambada, Hendrycks20MMLU, Clark19boolq, Zellers19Hellaswag, Clark18ARC, Cobbe21GSM8K, Wei22Cot}. 
All tasks utilize a 0-shot setting except for GSM8K, which uses 8-shot prompting.
Specifically, we evaluate Llama3 using the \texttt{gsm8k\_cot\_llama} task applying multi-turn chat templates and few-shot CoT prompting to reach its expected performance. In contrast, Qwen3 is evaluated on the standard \texttt{gsm8k\_cot} task with chat template disabled. 
For all other benchmarks, we omit chat templates. 
We report the average accuracy across all benchmarks excluding WikiText-2. 
Our end-to-end benchmarks additionally use IFEval~\cite{ifeval} (leaderboard settings) and set GSM8K to 0-shot. We furthermore use variable-length prompts from a processed ShareGPT dataset~\cite{sharegpt} to show the prefill/decoding scaling across different token ratios.

\begin{figure*}[t]
    \centering    
    \includegraphics[width=\textwidth]{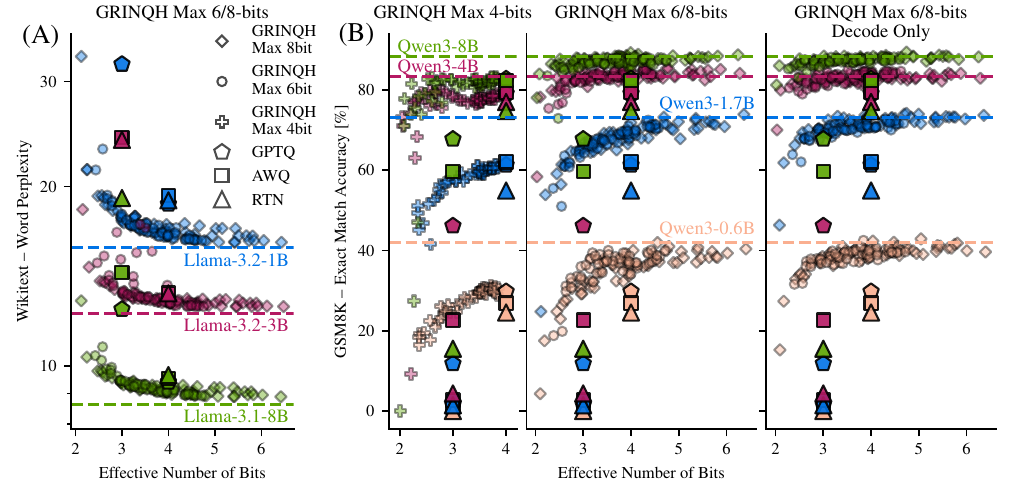}
    \caption{
    \textbf{GRINQH redefines the quantization Pareto frontier across model families and scales.} Dashed lines indicate BF16 baselines. 
    GRINQH precision distribution sweep is compared against iso-bit symmetric RTN, GPTQ, and AWQ baselines. 
    \textbf{(A)} WikiText-2 perplexity vs. effective bit width for the Llama3 Instruct family using $b_\text{max} \in \{6,8\}$. 
    \textbf{(B)} GSM8K CoT accuracy vs. effective bit width for the Qwen3 family. \textit{(Left panel)} GRINQH ($b_\text{max}=4$) sweep on GPTQ baseline.  
     \textit{(Middle panel)} GRINQH ($b_\text{max} \in \{6,8\}$ on RTN baseline).  
     \textit{(Right panel)} Real-case scenario with decoupled inference strategy ( $b_\text{max} \in \{6,8\}$ on RTN baseline) where dynamic precision loading is enabled exclusively for the memory-bound decoding stage.  
    }
    \label{fig:pareto_joint}
\end{figure*}

\textbf{Prefill vs. Decoding Evaluation.}\label{sec:prefillvsdecoding}
Standard benchmarks are predominantly prefill-heavy; therefore, to rigorously assess representational fidelity, we apply GRINQH's dynamic precision assignment to the prefill stage of all experiments performed with the reference implementation. Our kernel instead handles prefill by unpacking the weights at full ($b_\text{max}$) precision and performing dense matrix-matrix multiplication.
Additionally, we conduct specific evaluations on the decode-heavy GSM8K (CoT) 8-shot task to demonstrate GRINQH’s ability to decouple precision requirements between stages, effectively addressing the asymmetric computational demands of prefill and decoding (Fig.~\ref{fig:pareto_joint}B-decode-only, Fig.~\ref{fig:pareto_joint_appendix}B-decode-only). 

\textbf{Setup for Kernel Benchmarking.}
Our kernel is implemented in Triton~\cite{triton2019}. To obtain isolated decoding timings (see Fig.~\ref{fig:kernel_gemv_bench}), we utilize the framework's benchmarking suite with a standalone weight matrix of shape $W \in \mathbb{R}^{16{,}384 \times 4{,}096}$, representative of linear up-projection layers in modern 8B-parameter models.
We compare against baselines taken from \texttt{vLLM} (MARLIN~\cite{frantar24marlin}) and native PyTorch implementations (TorchAO~\cite{torchao}).
For our end-to-end benchmarks, we integrate GRINQH into \texttt{vLLM}, making it a selectable quantization method for any model given thresholds for the desired bit width.

\begin{table*}[t]
\centering
\caption{Benchmark results for the Llama3 8B and Qwen3 8B models. The average accuracy (Avg.) is calculated based on the accuracy metric of the following benchmarks: LAMBADA (LMB), ARC-C, HellaSwag, BoolQ, MMLU and GSM8K.}
\resizebox{\textwidth}{!}{
\begin{tabular}{ l | c c c c c | c c c c c } 
\toprule 
        & \multicolumn{5}{c|}{Llama-3.1-8B-Instruct} & \multicolumn{5}{c}{Qwen3-8B}  \\\cmidrule{2-11}
 Method & Eff.   & Wiki2  & LMB & GSM8K  & Avg. & Eff.   & Wiki2  & LMB  & GSM8K  & Avg. \\ 
        & Bits & ppl $\downarrow$  & ppl $\downarrow$  & acc $\uparrow$  & acc $\uparrow$ & Bits & ppl $\downarrow$  & ppl $\downarrow$  & acc $\uparrow$  & acc $\uparrow$ \\ 
\midrule 
\midrule 
BF16-Baseline                        & 16    & 8.64             & 3.40              & 85.14              & 74.35              & 16               & 12.20              & 4.59              & 88.32              & 74.02              \\\midrule
GPTQ~\cite{Frantar22}                & 4     & 9.41             & 3.64              & 80.82              & 72.85              & 4                & 12.68              & 5.17              & 82.79              & 72.18              \\
AWQ~\cite{Lin23}                     & 4     & 9.53             & 3.91              & 80.59              & 71.98              & 4                & 12.70              & 5.09              & 82.41              & 71.77              \\
RTN                                  & 4     & 9.67             & 3.90              & 79.83              & 71.44              & 4                & 12.89              & 5.40              & 74.75              & 69.89              \\
QuaRot+GPTQ~\cite{Ashkboos24}        & 4     & 9.38             & 3.55              & 81.88              & 72.86              & 4                & 12.58              & 4.81              & 86.28              & 72.99              \\ 
AutoRound~\cite{SignRoundV123}       & 4     & 9.62             & 3.89              & 80.89              & 72.25              & 4                & 13.30              & 5.94              & 84.08              & 71.79              \\ 
NVFP4-G16~\cite{abecassis_nvfp42025}                            & 4     & 9.27             & 3.41              & 82.49              & 73.24              & 4                & 12.38              & 5.07              & 86.13              & 73.36              \\
AnyPrecLLM~\cite{AnyPrecisionLLM24}  & 4     & 9.44             & 3.71              & 81.88              & 72.84              & 4                & 12.82              & 4.70              & 87.64              & 73.36              \\
SliM-LLM~\cite{SliM-LLM24}           & 4     & 9.35             & 3.46              & 82.64              & 72.82              & 4                & 12.61              & 5.20              & 87.41              & 72.50               \\
GRINQH-8b-RTN                        & 4.01  & \underline{8.97} & \underline{3.38}  & \textbf{85.14}     & \textbf{74.34}     & 3.97             & 12.29              & 4.58              & \underline{87.57} & 73.70               \\
GRINQH-8b-AnyP                       & 3.89  & \textbf{8.95}    & \textbf{3.36}     & \underline{84.53}  & \underline{74.19}  & 3.95             & \textbf{12.23}     & \textbf{4.44}     & \textbf{87.72}     & \textbf{73.91}     \\
GRINQH-6b-RTN                        & 3.92  & 9.02             & 3.41              & 84.31              & 73.91              & 3.94             & \underline{12.27}  & \underline{4.58}  & 87.19              & \underline{73.81}  \\\midrule
GPTQ~\cite{Frantar22}                & 3     & 12.46            & 5.86              & 60.35              & 63.27              & 3                & 14.89              & 6.96              & 67.78              & 64.67              \\
AWQ~\cite{Lin23}                     & 3     & 14.36            & 8.04              & 34.50              & 56.05              & 3                & 16.33              & 14.24             & 59.67              & 59.19              \\
RTN                                  & 3     & 19.13            & 7.43              & 6.60               & 50.64              & 3                & 22.84              & 41.29             & 15.54              & 44.79              \\
QuaRot+GPTQ~\cite{Ashkboos24}        & 3     & 12.80            & 5.77              & 57.54              & 63.40              & 3                & 14.28              & 7.30              & 68.84              & 65.03              \\ 
AutoRound~\cite{SignRoundV123}       & 3     & 11.81            & 5.24              & 49.58              & 63.00              & 3                & 15.08              & 7.31              & 67.78              & 65.45              \\ 
AnyPrecLLM~\cite{AnyPrecisionLLM24}  & 3     & 12.22            & 4.83              & 56.18              & 64.59              & 3                & 15.23              & 6.50              & 79.30              & 69.28              \\ 
SliM-LLM~\cite{SliM-LLM24}           & 3     & 12.01            & 4.85              & 68.46              & 66.35              & 3                & 14.88              & 5.58              & 77.63              & 67.64              \\
GRINQH-8b-RTN                        & 2.98  & \underline{9.30} & 3.49              & 81.50              & \underline{73.19}  & 3.07             & 12.61              & \underline{4.51}  & \textbf{88.02}     & \underline{73.33}  \\
GRINQH-8b-AnyP                       & 2.96  & \textbf{9.22}    & \textbf{3.37}     & \textbf{84.23}     & \textbf{73.69}     & 3.06             & \textbf{12.40}     & \textbf{4.35}     & \underline{87.41}  & \textbf{73.74}     \\ 
GRINQH-6b-RTN                        & 3.02  & 9.44             & \underline{3.40}  & \underline{82.41}  & 72.75              & 3.06             & \underline{12.58}  & 4.53              & 85.60              & 73.14              \\
GRINQH-4b-RTN                        & 3.02  & 9.80             & 3.80              & 79.68              & 71.34              & 3.03             & 13.02              & 5.30              & 79.38              & 70.94              \\
GRINQH-4b-GPTQ                       & 3.04  & 9.59             & 3.56              & 79.91              & 72.34              & 3.03             & 12.84              & 5.10              & 82.56              & 71.75              \\\midrule
AnyPrecLLM~\cite{AnyPrecisionLLM24}  & 2     & 1953             & 1e4             & 1.52               & 22.22              & 2                & 110                & 669               & 0.61               & 33.80              \\ 
SliM-LLM~\cite{SliM-LLM24}           & 2     & 535              & 3e4             & 2.96               & 19.59              & 2                & 101                & 212               & 2.12               & 31.44              \\
GRINQH-8b-RTN                        & 2.11  & \underline{12.87} & \underline{5.94} & \underline{53.53}  & \underline{61.37} & 2.25              & \underline{13.36}  & \underline{5.12}  & \underline{84.00}  & \underline{71.60}  \\
GRINQH-8b-AnyP                       & 2.01  & \textbf{10.47}   & \textbf{3.72}     & \textbf{71.65}     & \textbf{69.06}     & 2.22             & \textbf{12.95}     & \textbf{4.21}     & \textbf{87.04}     & \textbf{73.27}     \\ 
\bottomrule 
\end{tabular}
}
\label{tab:8B_comparison}
\end{table*}

\section{Results}
\label{sec:results}
\textbf{Quantization Performance and Pareto Frontiers.}
GRINQH demonstrates that near-lossless LLM inference is achievable at significantly reduced effective bit widths. 
As illustrated by the Pareto frontier for Qwen3 8B~(Fig.~\ref{fig:gsm8k_over_RTX4090}), our framework outperforms 4-bit state-of-the-art methods on GSM8K while operating 25\% faster on an RTX 4090 than optimized 4-bit MARLIN kernels. 
This speedup stems directly from our dynamic precision assignment, which enables the reduction or omission of non-critical weight channels. By prioritizing the protection of performance-critical weights, GRINQH drives down the average effective bit width without compromising model quality. 
Notably, GRINQH achieves up to $20\times$ lower accuracy degradation compared to competitive 3-bit methods, effectively reconciling hardware efficiency with representational integrity.

\textbf{Pareto Frontier Across Scales.} To demonstrate GRINQH's robustness across model scales, Fig.~\ref{fig:pareto_joint} illustrates the emergence of a new SOTA Pareto frontier spanning the 2--6 bit range. 
For larger models such as Qwen3 4B and 8B (Fig.~\ref{fig:pareto_joint}B (middle)) performance tracks the BF16 baseline closely down to 2.5 bits when using $b_\text{max} \in \{6,8\}$, effectively closing the accuracy gap that is traditionally associated with 3-bit quantization. 
Similarly, for the Llama3 family (Fig.~\ref{fig:pareto_joint}A), GRINQH maintains effective performance deep into the 2--3 bit regime, preventing the catastrophic perplexity explosions characteristic of low-bit quantization. In this regime, GRINQH significantly separates itself from widely adopted methods like GPTQ and AWQ at 3 bits.
Across all evaluated scales and architectures, our framework consistently enables high-fidelity performance at fractional bit widths, establishing a superior Pareto frontier for low-bit LLM inference. Further results are provided in Appendix~\ref{app:paretoFrontier}.

\textbf{The GRINQH max-4-bit Baseline.} 
Even when restricted to a 4-bit base ($b_\text{max}=4$), GRINQH establishes a superior Pareto frontier (Fig.~\ref{fig:pareto_joint}B (left panel)).
The GRINQH data points seamlessly emerge from the 4-bit-baseline, with large Qwen3 models maintaining near-lossless performance for a significant range of bit width before declining.
While GRINQH with $b_\text{max}=4$ cannot exceed its own 4-bit source, it consistently outperforms specialized 3-bit RTN, AWQ and GPTQ models.  
This suggests that GRINQH's dynamic bit-planar fetching is a more robust mechanism for bit-rate reduction than static low-bit calibration. At the same time, it highlights that GRINQH can also build on top of sophisticated quantization methods with any quantized base precision format. Consequently, GRINQH enables to seamlessly scale down effective precision, achieving accelerated inference without further architectural changes. 

\textbf{Cross-Benchmark Superiority.} 
Tab.~\ref{tab:8B_comparison} compares Llama3 8B and Qwen3 8B across the benchmark suite (full results for all scales are in Appendix~\ref{app:extendedBenchmarks}).
At an effective bit width of $\sim4$ bits ($b_\text{max} \in \{6,8\}$), GRINQH matches the BF16 baseline and consistently outperforms all uniform and non-uniform approaches. This stability extends into the sub-3-bit regime, where static PTQ typically suffers from catastrophic representational collapse. \\
Additionally, GRINQH pushed to an effective width of 2.01--2.22 bits outperforms all tested SOTA methods operating at a full 3 bits, saving 1 bit per weight while simultaneously delivering superior accuracy. Notably, for Qwen3 8B, our 2.22-bit configuration even surpasses the average accuracy of the majority of 4-bit baselines. While competitive methods (exemplified by any-precision LLM and SliM-LLM) collapse toward chance-level accuracy at 2 bits, GRINQH maintains robustness. Although utilizing nested weights (any-precision LLM) provides a performance boost over simple bit-slicing in this ultra-low-bit regime, the fact that our RTN-based configuration also maintains high accuracy demonstrates that GRINQH’s superiority is driven primarily by its inherent precision allocation logic rather than the underlying quantization primitive. 

\begin{figure}[t]
  \begin{minipage}[c]{0.59\textwidth}
    \vspace{0pt} 
    \centering
    \captionof{table}{Tok/s performance over different effective bit widths and input:output ratios (prefill: (3000:1), decoding (1:3000)) of tokens on random data. Relative performance is calculated against GPTQ 4-bit MARLIN.}
\label{tab:vllm_prefill}
\resizebox{\textwidth}{!}{
    \begin{tabular}{lcccc}
    \toprule
    \textbf{Kernel} & \textbf{Prefill}  & \textbf{Decoding} \\
                    & \textbf{Abs. (\,Rel.\,)}  & \textbf{Abs. (\,Rel.\,)} \\
    \midrule
    Eff. 2-bit GRINQH-8b & \multirow{3}{*}{8187 (\,0.80$\times$\,)} & 204.6 (\,1.28$\times$\,) \\
    Eff. 3-bit GRINQH-8b &                                  & 182.2 (\,1.14$\times$\,) \\
    Eff. 4-bit GRINQH-8b &                                  & 157.6 (\,0.98$\times$\,) \\
    \midrule
    Eff. 4-bit GRINQH-6b & 8241 (\,0.81$\times$\,) & 165.1 (\,1.03$\times$\,) \\
    Eff. 3-bit GRINQH-4b & 8241 (\,0.81$\times$\,) & 165.1 (\,1.03$\times$\,) \\
    \midrule
    GPTQ 4-bit MARLIN & 10170 (\,1.00$\times$\,) & 160.4 (\,1.00$\times$\,) \\
    RTN 8-bit MARLIN  &  8927 (\,0.88$\times$\,) & 100.2 (\,0.62$\times$\,) \\
    \bottomrule
    \end{tabular}
}
  \end{minipage}\hfill
  \begin{minipage}[c]{0.40\textwidth}
    \vspace{0pt} 
    \centering
    \includegraphics[width=0.96\textwidth]{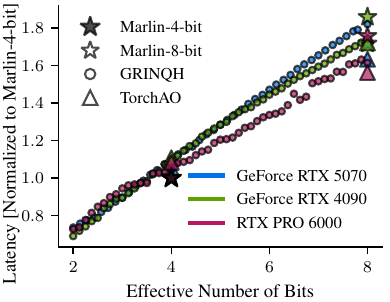}
    \caption{
      Normalized isolated kernel runtimes over a range of target effective bit widths. Times are normalized w.r.t. 4-bit Marlin kernel on the same device. 
    }
    \label{fig:kernel_gemv_bench}
  \end{minipage}
\end{figure}
\begin{figure*}[t]
    \centering    
    \includegraphics[width=\textwidth]{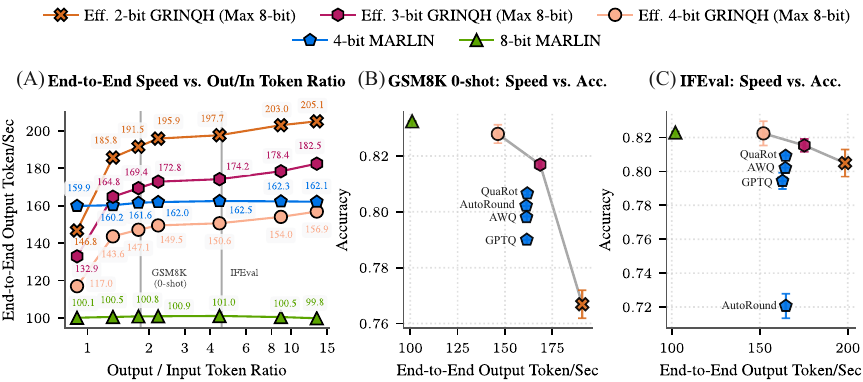}
    \caption{End-to-End performance scaling and task evaluation. \textbf{(A)} Decoding throughput (tokens/s) over increasingly longer outputs relative to prompts selected from the ShareGPT dataset, averaged over 1000 samples across two seeds ($~225$ avg. input tokens per prompt). While prefill initially weighs down the average throughput, GRINQH delivers SOTA performance even on tasks with considerable relative prefill. Vertical lines represent the average out/in ratio for GSM8K 0-shot (B) and IFEval (C) respectively. MARLIN 4-bit evaluated on GPTQ, 8-bit on RTN. \textbf{(B, C)} Speed vs. strict accuracy for GSM8K 0-shot with 4096 maximum output tokens per prompt and IFEval (leaderboard settings) with 256.  In the low-bit regime, GRINQH consistently forms the Pareto front. Accuracy refers to ``strict-match'' for GSM8K and ``inst\_level\_strict\_acc'' for IFEval. All measurements for NVIDIA RTX 4090 using Llama3.1-8B.
    }
    \label{fig:vllm_integration}
\end{figure*}
\textbf{Representational Fidelity in Decoupled Inference.}
We evaluate GRINQH in a realistic inference setting where prefill and decoding are treated separately.
Since the prefill stage is primarily compute-bound, loading weights at $b_\text{max}$ introduces negligible latency overhead. We therefore maintain maximum precision during prefill to preserve representational fidelity, while employing GRINQH’s dynamic mixed-precision logic exclusively during the memory-bound decoding stage. 
Fig.~\ref{fig:pareto_joint}B (right panel) shows that this stage-aware approach yields a significant performance boost;  
the resulting Pareto frontiers are more densely packed and exhibit a delayed accuracy drop compared to the stage-agnostic counterparts. This enables even smaller models to operate effectively below an average of 3 bits, demonstrating that by decoupling these stages, GRINQH bypasses the traditional accuracy-efficiency trade-off and protects initial context processing without sacrificing decoding throughput.

\textbf{Decoding Throughput and Bit-Width Scaling.}
To evaluate hardware efficiency, we benchmark our decoding kernel against state-of-the-art baselines. Our implementation achieves performance parity with optimized 4-bit and 8-bit kernels while exhibiting graceful linear scaling across intermediate effective precisions (Fig.~\ref{fig:kernel_gemv_bench}). This confirms that the kernel’s runtime scales proportionally with the effective bit width, showing that non-uniform precision distributions incur no relevant overhead. We validate these results across a range of architectures, including consumer-grade (NVIDIA RTX 4090, NVIDIA RTX 5070) and professional-grade (NVIDIA RTX PRO 6000) GPUs. 

\textbf{End-to-end Measurements.}
For our kernel we perform $b_\text{max}$ dense prefill. Due to efficient unpacking and kernel fusion this is only 20\% slower than optimized prefill for 4-bit weights, see Tab.~\ref{tab:vllm_prefill}. Limiting the maximum bits to $<8$ allows for lower sustained memory usage and gives a slight boost to prefill. The slower prefill performance is outweighed by up to 28\% faster decoding speed. 
As shown in Fig.~\ref{fig:vllm_integration}A, the performance gains observed in both isolated kernel (Fig.~\ref{fig:kernel_gemv_bench}) and full-model (Tab.~\ref{tab:vllm_prefill}) benchmarks translate well to end-to-end speedups (measured in output tokens/s) over the full inference duration, including prefill. This holds for all evaluated tasks, establishing Pareto-optimal points for all GRINQH configurations runs in Fig.~\ref{fig:vllm_integration}B,C. Notably, our 3-bit effective configuration retains significantly higher accuracy than existing 4-bit SOTA methods while simultaneously offering a superior end-to-end speedup. This result effectively demonstrates that GRINQH can surpass the accuracy-efficiency trade-off of static 4-bit quantization, redefining the performance ceiling for low-bit LLM deployment. 
\section{Conclusion}
\label{sec:conclusion}
In this work, we propose GRINQH, a decoding-centric quantization framework that addresses the memory-bandwidth bottleneck in LLM inference through input-dependent precision assignment. 
By unifying dynamic sparsity and quantization, GRINQH substantially reduces the volume of data transferred from off-chip memory during decoding. 
This fine-grained control of precision enables high-throughput inference while surpassing the accuracy of state-of-the-art quantization methods at equivalent, or even significantly lower effective bit widths.

GRINQH functions as a quantization-agnostic framework that enhances existing formats like RTN or GPTQ, while further benefiting from sophisticated nested weight techniques such as any-precision LLM in the sub-3-bit regime.
Notably, the inherent structure of these nested weights provides elastic multi-precision profiles within a single set of parameters, offering a promising architectural foundation for future research into speculative decoding and adaptive precision scheduling.

Yet, GRINQH targets batch-size-one scenarios and utilizes bit-planar storage, which necessitates a $b_\text{max}$-DRAM footprint. However, when restricted to a static 4-bit memory footprint ($b_\text{max}=4$), GRINQH's 3-bit inference closely matches the accuracy of 4-bit SOTA baselines for some models while significantly reducing DRAM-to-SRAM traffic. Furthermore, while prefill performance is slightly impacted by the requirement to unpack bit-planar representations, this is balanced by the advantages of stage-aware optimization. By decoupling inference stages, GRINQH achieves near-lossless context processing during prefill while leveraging dynamic precision during decoding to protect performance-critical outliers and adapt to activation distribution shifts.
Additionally, GRINQH unlocks flexible bit widths that more fluidly adapt to model and hardware needs, unconstrained by standard fixed-precision grids. Ultimately, GRINQH translates theoretical gains into tangible reductions in memory traffic, redefining the Pareto frontier for low-bit LLM deployment.

\begin{ack}
This work was sponsored by the Federal Ministry of Education, Germany BMBF under project NEUROTEC-II grants no. 16ME0398K and 16ME0399, and 01IS22094E WestAI - AI Service Center West; and Neurosys as part of the initiative "Cluster4Future" funded by the Federal Ministry of Education and Research BMBF (03ZU1106CB); and Phase II: NeuroSys as part of the initiative ``Clusters4Future'' funded by the Federal Ministry of Research, Technology and Space BMFTR (03ZU2106CB). The authors gratefully acknowledge computing time on the supercomputer JURECA~\cite{jureca2021} and JUWELS~\cite{alvarez2021juwels} at Forschungszentrum Jülich.
They further thank Abigail Morrison for support during this work, and Viet Anh Khoa Tran and Matthias Oberländer for providing thoughtful comments on the manuscript.
\end{ack}

\section*{Author Contributions}

JO helped define the trajectory of the GRINQH method, performed the empirical evaluation, including design of model calibration and quantization, accuracy experiments, and the implementation of baseline comparisons. JO implemented the initial functional prototype within vLLM to validate the empirical proof of concept and led the manuscript preparation. JF conceptualized the GRINQH approach, developed the main algorithmic formulation, designed and implemented the custom Triton decoding kernel, contributed to the prefill implementation strategy, and supervised the technical development and evaluation strategy of the project. CS developed the PyTorch-based prefill path, contributed to the Triton kernel development, and led the benchmarking, profiling, and final kernel integration into vLLM. EN provided project supervision and guidance. All authors contributed to the interpretation of results and the final revision of the manuscript.

\bibliography{Literatur}

\newpage
\FloatBarrier
\appendix

\section{Appendix: Hyperparameter Selection and Tuning} \label{app:HyperparameterSelection}

This section details the formalization of our precision distribution (Section~\ref{app:precisionDistributionFormalization}), describes the random sweep used to validate the robustness of the framework (Section~\ref{app:DistributionSweep}), and provides general recommendations for selecting suitable hyperparameters (Section~\ref{app:HyperparameterGuide}).

A central finding of this work is that GRINQH consistently outperforms state-of-the-art (SOTA) fixed-bit baselines across a wide range of precision allocations. This suggests that the performance gains are inherent to our dynamic execution layer rather than a result of exhaustive hyperparameter over-tuning. While we formulate the optimal distribution as a constrained optimization problem, we leave the development of automated search strategies or a training-based approach for future work.

\subsection{Precision Distribution Formalization} \label{app:precisionDistributionFormalization}

A core hyperparameter in our framework is the \textbf{precision distribution}, which defines the fraction of a model's weights assigned to specific precision levels within a linear layer. 
Specifically, given a maximum bit width $b_{\text{max}}$, we define a set of available precisions $\mathcal{B} = \left(b_0, b_1, \dots, b_n\right)$, where $b_i = 2i$ and $n = b_{\text{max}}/2$.
The precision distribution is characterized by a vector of fractions $\mathrm{P} = \left(p_0,p_1,\dots, p_n\right) \in \Delta^{n+1}$, where $\Delta^{n+1}$ denotes the probability simplex such that $\sum p_i = 1$ and $p_i \geq 0$.
Each $p_i$ represents the portion of weights allocated to bit width $b_i$.

To ensure the resulting model meets the user-defined target effective bit width ($b^*$), the distribution is constrained such that the weighted average bit width approximates $b^*$. We formalize the selection of $P$ as a constrained optimization problem:
\begin{equation}
\begin{aligned}
\max_{P} \quad \mathcal{M}(\text{Model}(P)) \quad
\text{s.t.} \quad & \sum_{i=0}^n p_i = 1, \quad p_i \geq 0,
& \left| \left( \sum_{i=0}^n p_i \cdot b_i \right) - b^* \right| \leq \epsilon
\end{aligned}
\end{equation}
where $\mathcal{M}$ represents a (task-specific) performance metric (e.g., accuracy or negative perplexity) and
$\epsilon$ is a small tolerance parameter (e.g., $0.01$) introduced to allow for a wider range of valid precision distributions.

\paragraph{Implementation via GRINQH.} 
While the optimization problem determines the ideal configuration $P$, the GRINQH framework realizes the precision allocation of the weights by applying thresholds to the magnitudes of their corresponding activations. These thresholds are calibrated based on $P$.
Specifically, for every calibration step, we identify the separating magnitudes (thresholds) that partition the activations into the fractions defined by $P$ by applying a top-$p$ approach. For each linear layer $\ell$, these values are logged and subsequently averaged across all forward passes to obtain the final vector of fixed thresholds:
\begin{align}
\Theta^{(\ell)} = \left( \theta_1^{(\ell)}, \dots, \theta_{n-1}^{(\ell)}\right)
\end{align}
Fig.~\ref{fig:kernel_aux_data_bench}B shows that these calibrated thresholds successfully maintain the target effective bit width $b^*$ during inference across a variety of benchmark tasks. 
A detailed robustness analysis of this calibration phase is provided in Appendix~\ref{app:calibration_analysis}.

\subsection{Search Space and Robustness}\label{app:DistributionSweep}
To generate the Pareto curves (Fig.~\ref{fig:gsm8k_over_RTX4090}, Fig.~\ref{fig:pareto_joint}, Fig.~\ref{fig:pareto_joint_appendix}), we conducted a random sweep over the configuration space. The sampled distributions are detailed in Tab.~\ref{tab:precision_distribution_sweep}. These distributions were selected to ensure a representative spread across effective bit widths $b^* \in [2,7]$.

It is important to note that in this work we did not explicitly solve the optimization problem defined in the previous section. Instead, we sampled valid precision distributions $\mathrm{P}$ that satisfy the bit-width constraints without applying any sophisticated methods to maximize the performance metric $\mathcal{M}$. We observe that even with these randomly sampled configurations, the majority of our results consistently outperform state-of-the-art (SOTA) fixed-bit baselines. This suggests that the framework is highly robust to the specific choice of $\mathrm{P}$ and that accuracy performance gains are inherent to the flexible precision allocation rather than hyperparameter over-tuning.

As shown by the Pareto frontiers, performance is significantly more tied to the target bit width $b^*$ than to the specific distribution used, particularly in the lower-bit regime. However, the observed variance between different configurations at identical bit widths suggests that solving the optimization problem for $P$ remains a promising direction for future research. In the following section, we provide a heuristic guide for identifying ``good'' hyperparameter sets that yield strong performance without requiring an exhaustive search.

\begingroup
\small
\renewcommand{\arraystretch}{1.2}

\begin{longtable}{c | p{0.3\textwidth} | p{0.3\textwidth} | p{0.25\textwidth}}

\caption{Precision distribution (P) sweep for various $b_\text{max}$ values, where ($p_0, p_1, p_2, p_3, p_4$) represents the fractions of 0, 2, 4, 6, and 8-bit widths, respectively. The distributions are sampled with the constraint $0.15 \leq p_0 \leq 0.30$, ensuring a significant level of activation sparsity. To maintain focus on the low-to-mid bit-width regime, the 8-bit fraction ($p_4$) is constrained to 0.0 for most samples. All distributions are filtered to ensure the resulting effective bit width ($b$) satisfies $2 \leq b \leq 7$.}
\label{tab:precision_distribution_sweep} \\

\hline
\# & $b_\text{max}=8$ \newline ($p_0, p_1, p_2, p_3, p_4$) & $b_\text{max}=6$ \newline ($p_0, p_1, p_2, p_3$) & $b_\text{max}=4$ \newline ($p_0, p_1, p_2$) \\
\hline
\endfirsthead

\hline
\# & $b_\text{max}=8$ \newline ($p_0, p_1, p_2, p_3, p_4$) & $b_\text{max}=6$ \newline ($p_0, p_1, p_2, p_3$) & $b_\text{max}=4$ \newline ($p_0, p_1, p_2$) \\
\hline
\endhead
1  & 0.30\quad0.35\quad0.28\quad0.07\quad0.00 & 0.30\quad0.35\quad0.18\quad0.17 & 0.00\quad0.00\quad1.00 \\
2  & 0.15\quad0.68\quad0.13\quad0.04\quad0.00 & 0.15\quad0.56\quad0.13\quad0.16 & 0.20\quad0.05\quad0.75 \\
3  & 0.25\quad0.30\quad0.41\quad0.04\quad0.00 & 0.25\quad0.30\quad0.30\quad0.15 & 0.18\quad0.04\quad0.78 \\
4  & 0.20\quad0.40\quad0.32\quad0.08\quad0.00 & 0.20\quad0.36\quad0.28\quad0.16 & 0.15\quad0.05\quad0.80 \\
5  & 0.30\quad0.21\quad0.33\quad0.16\quad0.00 & 0.30\quad0.21\quad0.33\quad0.16 & 0.25\quad0.03\quad0.72 \\
6  & 0.35\quad0.07\quad0.51\quad0.07\quad0.00 & 0.35\quad0.07\quad0.41\quad0.17 & 0.30\quad0.02\quad0.68 \\
7  & 0.20\quad0.32\quad0.44\quad0.04\quad0.00 & 0.20\quad0.32\quad0.30\quad0.18 & 0.05\quad0.15\quad0.80 \\
8  & 0.30\quad0.17\quad0.36\quad0.17\quad0.00 & 0.30\quad0.17\quad0.36\quad0.17 & 0.04\quad0.18\quad0.78 \\
9  & 0.25\quad0.22\quad0.31\quad0.22\quad0.00 & 0.25\quad0.22\quad0.31\quad0.22 & 0.03\quad0.17\quad0.80 \\
10 & 0.25\quad0.19\quad0.37\quad0.19\quad0.00 & 0.25\quad0.19\quad0.37\quad0.19 & 0.02\quad0.20\quad0.78 \\
11 & 0.25\quad0.08\quad0.59\quad0.08\quad0.00 & 0.25\quad0.08\quad0.49\quad0.18 & 0.01\quad0.19\quad0.80 \\
12 & 0.25\quad0.25\quad0.24\quad0.26\quad0.00 & 0.25\quad0.25\quad0.24\quad0.26 & 0.10\quad0.05\quad0.85 \\
13 & 0.20\quad0.16\quad0.56\quad0.08\quad0.00 & 0.20\quad0.16\quad0.46\quad0.18 & 0.08\quad0.07\quad0.85 \\
14 & 0.20\quad0.24\quad0.32\quad0.24\quad0.00 & 0.20\quad0.24\quad0.32\quad0.24 & 0.06\quad0.09\quad0.85 \\
15 & 0.20\quad0.20\quad0.40\quad0.20\quad0.00 & 0.20\quad0.20\quad0.40\quad0.20 & 0.05\quad0.10\quad0.85 \\
16 & 0.15\quad0.13\quad0.68\quad0.04\quad0.00 & 0.15\quad0.13\quad0.50\quad0.22 & 0.04\quad0.11\quad0.85 \\
17 & 0.30\quad0.15\quad0.16\quad0.39\quad0.00 & 0.30\quad0.15\quad0.16\quad0.39 & 0.12\quad0.03\quad0.85 \\
18 & 0.25\quad0.10\quad0.38\quad0.20\quad0.07 & 0.25\quad0.10\quad0.38\quad0.27 & 0.10\quad0.04\quad0.86 \\
19 & 0.15\quad0.21\quad0.43\quad0.21\quad0.00 & 0.15\quad0.21\quad0.43\quad0.21 & 0.08\quad0.05\quad0.87 \\
20 & 0.15\quad0.09\quad0.67\quad0.09\quad0.00 & 0.15\quad0.09\quad0.47\quad0.29 & 0.06\quad0.06\quad0.88 \\
21 & 0.30\quad0.05\quad0.32\quad0.26\quad0.07 & 0.30\quad0.05\quad0.32\quad0.33 & 0.05\quad0.05\quad0.90 \\
22 & 0.30\quad0.05\quad0.26\quad0.36\quad0.03 & 0.30\quad0.05\quad0.26\quad0.39 & 0.02\quad0.08\quad0.90 \\
23 & 0.25\quad0.05\quad0.35\quad0.28\quad0.07 & 0.25\quad0.05\quad0.35\quad0.35 & 0.03\quad0.07\quad0.90 \\
24 & 0.20\quad0.18\quad0.17\quad0.45\quad0.00 & 0.20\quad0.18\quad0.17\quad0.45 & 0.04\quad0.06\quad0.90 \\
25 & 0.25\quad0.05\quad0.40\quad0.22\quad0.08 & 0.25\quad0.05\quad0.40\quad0.30 & 0.05\quad0.04\quad0.91 \\
26 & 0.25\quad0.08\quad0.15\quad0.52\quad0.00 & 0.25\quad0.08\quad0.15\quad0.52 & 0.01\quad0.09\quad0.90 \\
27 & 0.15\quad0.19\quad0.18\quad0.48\quad0.00 & 0.15\quad0.19\quad0.18\quad0.48 & 0.15\quad0.02\quad0.83 \\
28 & 0.00\quad0.00\quad1.00\quad0.00\quad0.00 & 0.00\quad0.00\quad1.00\quad0.00 & 0.18\quad0.02\quad0.80 \\
29 & 0.20\quad0.08\quad0.24\quad0.48\quad0.00 & 0.20\quad0.08\quad0.24\quad0.48 & 0.22\quad0.01\quad0.77 \\
30 & 0.15\quad0.05\quad0.48\quad0.24\quad0.08 & 0.15\quad0.05\quad0.48\quad0.32 & 0.25\quad0.01\quad0.74 \\
31 & 0.20\quad0.16\quad0.08\quad0.56\quad0.00 & 0.20\quad0.16\quad0.08\quad0.56 & 0.28\quad0.01\quad0.71 \\
32 & 0.25\quad0.10\quad0.06\quad0.39\quad0.20 & 0.25\quad0.10\quad0.06\quad0.59 & 0.03\quad0.03\quad0.94 \\
33 & 0.30\quad0.06\quad0.07\quad0.50\quad0.07 & 0.30\quad0.06\quad0.07\quad0.57 & 0.02\quad0.04\quad0.94 \\
34 & 0.15\quad0.09\quad0.17\quad0.59\quad0.00 & 0.15\quad0.09\quad0.17\quad0.59 & 0.01\quad0.05\quad0.94 \\
35 & 0.25\quad0.05\quad0.07\quad0.42\quad0.21 & 0.25\quad0.05\quad0.07\quad0.63 & 0.04\quad0.02\quad0.94 \\
36 & 0.20\quad0.06\quad0.06\quad0.64\quad0.04 & 0.20\quad0.06\quad0.06\quad0.68 & 0.05\quad0.01\quad0.94 \\
37 & 0.20\quad0.10\quad0.10\quad0.42\quad0.18 & 0.20\quad0.10\quad0.28\quad0.42 & 0.25\quad0.10\quad0.65 \\
38 & 0.30\quad0.05\quad0.05\quad0.21\quad0.39 & 0.30\quad0.05\quad0.44\quad0.21 & 0.30\quad0.05\quad0.65 \\
39 & 0.15\quad0.10\quad0.10\quad0.50\quad0.15 & 0.15\quad0.25\quad0.10\quad0.50 & 0.20\quad0.20\quad0.60 \\
40 & 0.30\quad0.05\quad0.30\quad0.20\quad0.15 & 0.30\quad0.05\quad0.30\quad0.35 & 0.15\quad0.25\quad0.60 \\
41 & 0.25\quad0.15\quad0.10\quad0.38\quad0.12 & 0.25\quad0.15\quad0.10\quad0.50 & 0.25\quad0.20\quad0.55 \\
42 & 0.30\quad0.09\quad0.14\quad0.07\quad0.40 & 0.30\quad0.09\quad0.14\quad0.47 & 0.25\quad0.25\quad0.50 \\
43 & 0.15\quad0.08\quad0.13\quad0.60\quad0.04 & 0.15\quad0.08\quad0.13\quad0.64 & 0.30\quad0.20\quad0.50 \\
44 & 0.15\quad0.05\quad0.12\quad0.59\quad0.09 & 0.15\quad0.05\quad0.12\quad0.68 & 0.35\quad0.10\quad0.55 \\
45 & 0.25\quad0.00\quad0.00\quad0.52\quad0.23 & 0.25\quad0.00\quad0.00\quad0.75 & 0.15\quad0.40\quad0.45 \\
46 & 0.30\quad0.00\quad0.00\quad0.28\quad0.42 & 0.20\quad0.00\quad0.10\quad0.70 & 0.25\quad0.30\quad0.45 \\
47 & 0.15\quad0.00\quad0.25\quad0.34\quad0.26 & 0.15\quad0.00\quad0.25\quad0.60 & 0.25\quad0.34\quad0.41 \\
48 & 0.30\quad0.00\quad0.00\quad0.21\quad0.49 & 0.30\quad0.00\quad0.00\quad0.70 & 0.27\quad0.33\quad0.40 \\
49 & 0.15\quad0.00\quad0.00\quad0.68\quad0.17 & 0.15\quad0.18\quad0.17\quad0.50 & 0.30\quad0.33\quad0.37 \\
50 & 0.15\quad0.00\quad0.08\quad0.51\quad0.26 & 0.15\quad0.00\quad0.34\quad0.51 & 0.30\quad0.28\quad0.42 \\
51 & 0.30\quad0.00\quad0.00\quad0.07\quad0.63 & 0.30\quad0.00\quad0.07\quad0.63 & 0.35\quad0.23\quad0.42 \\
52 & 0.25\quad0.00\quad0.00\quad0.23\quad0.52 & 0.25\quad0.00\quad0.23\quad0.52 & 0.35\quad0.25\quad0.40 \\
53 & 0.25\quad0.00\quad0.00\quad0.15\quad0.60 & 0.25\quad0.00\quad0.15\quad0.60 &  \\
54 & 0.20\quad0.00\quad0.00\quad0.32\quad0.48 & 0.20\quad0.00\quad0.32\quad0.48 &  \\
55 & 0.25\quad0.00\quad0.00\quad0.08\quad0.67 & 0.25\quad0.08\quad0.00\quad0.67 &  \\
56 & 0.20\quad0.00\quad0.00\quad0.26\quad0.54 & 0.20\quad0.13\quad0.13\quad0.54 &  \\
57 & 0.00\quad0.00\quad0.00\quad1.00\quad0.00 & 0.00\quad0.00\quad0.00\quad1.00 &  \\
58 & 0.20\quad0.00\quad0.00\quad0.16\quad0.64 & 0.20\quad0.00\quad0.16\quad0.64 &  \\
59 & 0.15\quad0.00\quad0.00\quad0.17\quad0.68 & 0.15\quad0.00\quad0.17\quad0.68 &  \\
60 & 0.00\quad0.00\quad0.00\quad0.00\quad1.00 & 0.15\quad0.25\quad0.40\quad0.20 &  \\
\hline
\end{longtable}
\endgroup

\paragraph{Table Model Selection Criteria.}
For the table comparisons in the main text and appendix (Tab.~\ref{tab:8B_comparison}, Tab.~\ref{tab:qwen-comparison-appendix}, Tab.~\ref{tab:llama-comparison-appendix}), we selected configurations based on their Wikitext-2 perplexity. In this context, Wikitext-2 perplexity serves as our proxy for the performance metric $\mathcal{M}$ that is aimed to be maximized. This selection criterion demonstrates that configurations chosen to optimize for a general language modeling objective generalize robustly to specific downstream accuracy-based tasks.

\subsection{Empirical Guidelines for Configuration} \label{app:HyperparameterGuide}

\begin{figure*}[t]
    \centering    
    \includegraphics[width=\textwidth]{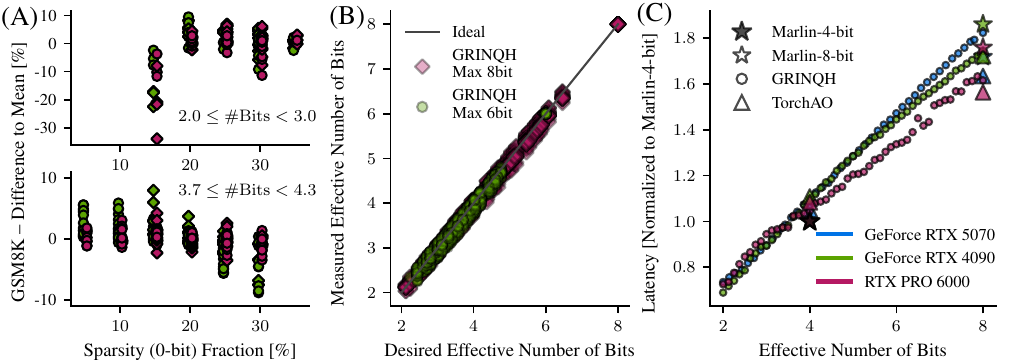}
    \caption{
    \textbf{(A)} Sensitivity analysis of the 0-bit (sparsity) fraction ($p_0$) for specific bit-width windows (Top: 2.0--3.0 bits; Bottom: 3.7--4.3 bits). Data points indicate the relative GSM8K performance change across varying sparsity levels with respect to the average performance in that window. Results show that high sparsity benefits low-bit regimes but penalizes higher bit regimes, suggesting that optimal $p_0$ configuration is bit width dependent.
    \textbf{(B)} Comparison between the desired effective bit width and the measured effective bits. The GRINQH data points are obtained over all benchmark tasks and various models ($b_\text{max} \in \{6, 8\}$).
     The close adherence to the identity mapping (black line) confirms that the calibrated thresholds accurately achieve target bit-rates across diverse model distributions.   
      \textbf{(C)} Kernel performance comparison on NVIDIA GPUs. Runtimes are measured in an isolated kernel setup over a range $[2,8]$ of target effective bit widths.
      Baselines use their respective state-of-the-art native kernels, which the GRINQH trend line closely rivals around the fixed-width points. For each device, latency numbers are normalized with respect to the best performing 4-bit MARLIN kernel run on the same device, measuring respectively $[0.072, 0.057, 0.034]$ms on the RTX5070/RTX4090/RTX PRO 6000.
    }
    \label{fig:kernel_aux_data_bench}
\end{figure*}

The flexibility of the GRINQH framework allows for a vast configuration space. Based on our sensitivity analysis, we provide the following observations and a streamlined workflow for identifying high-performance precision distributions.

\paragraph{Regime-Dependent Sparsity Trade-offs.}
Our analysis (Fig.~\ref{fig:kernel_aux_data_bench}A) reveals that the relationship between sparsity and performance is highly dependent on the target bit-budget. In extreme low-bit regimes ($2.0 \leq b \leq 3.0$), higher sparsity levels ($\geq 20\%$) act as a critical enabler. By aggressively exploiting ``free sparsity'' to prune non-salient channels (0-bit allocation), GRINQH reallocates the saved budget to high-precision outliers. Without this strategic reallocation, the model lacks the bit-depth necessary to preserve critical activation saliency. 

Conversely, in moderate bit-regimes ($3.7 \leq b \leq 4.3$), the budget is inherently sufficient for outlier protection. In these cases, exceeding the ``free sparsity'' threshold ($\sim 20\%$) can degrade performance by omitting necessary weight information. Thus, GRINQH’s primary strength lies in its ability to navigate these regimes dynamically, operating within the free sparsity window by default while strategically exceeding it when bit-constraints become severe.

\begin{figure*}[ht]
    \centering    
    \includegraphics[width=0.7\textwidth]{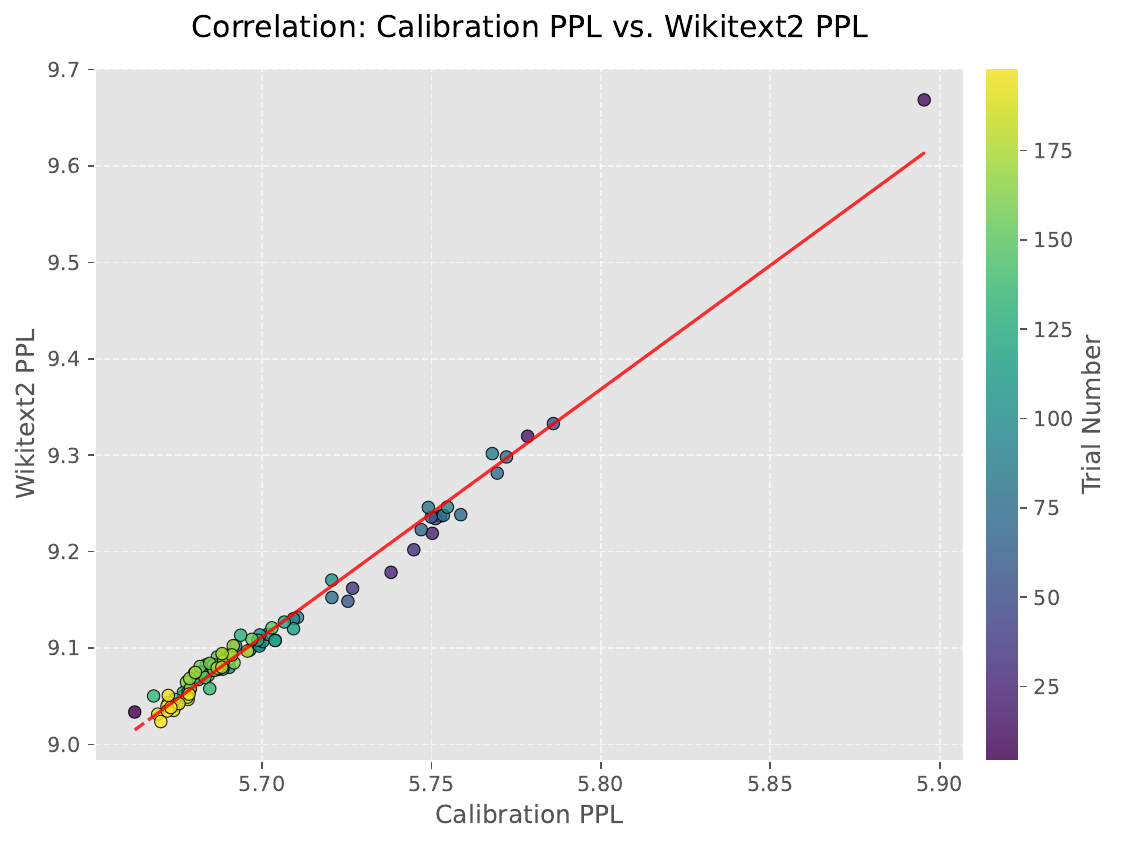}
    \caption{
    \textbf{Correlation between calibration and downstream perplexity.   
    }
    Each data point represents a unique precision distribution configuration $\mathrm{P}$ sampled during our hyperparameter sweep. Calibration PPL is computed on The Pile (Uncopyrighted), while downstream PPL is evaluated on Wikitext-2. The strong linear correlation validates calibration PPL as a reliable and computationally efficient proxy metric for performance optimization during threshold selection.
    }
    \label{fig:correlationCalibrationWikitext}
\end{figure*}

\paragraph{Search Space Selection and Proxy Metrics}
To identify optimal fractions $p_i$, we utilize calibration perplexity (PPL) as a proxy metric $\mathcal{M}$. Fig.~\ref{fig:correlationCalibrationWikitext} shows that calibration PPL correlates strongly with downstream Wikitext2 PPL, which in turn serves as a reliable indicator of task-specific accuracy (see Section~\ref{app:DistributionSweep}). Using calibration PPL as a signal is computationally efficient, as it is obtained by the existing calibration pipeline.

\begin{figure*}[ht]
    \centering    
    \includegraphics[width=\textwidth]{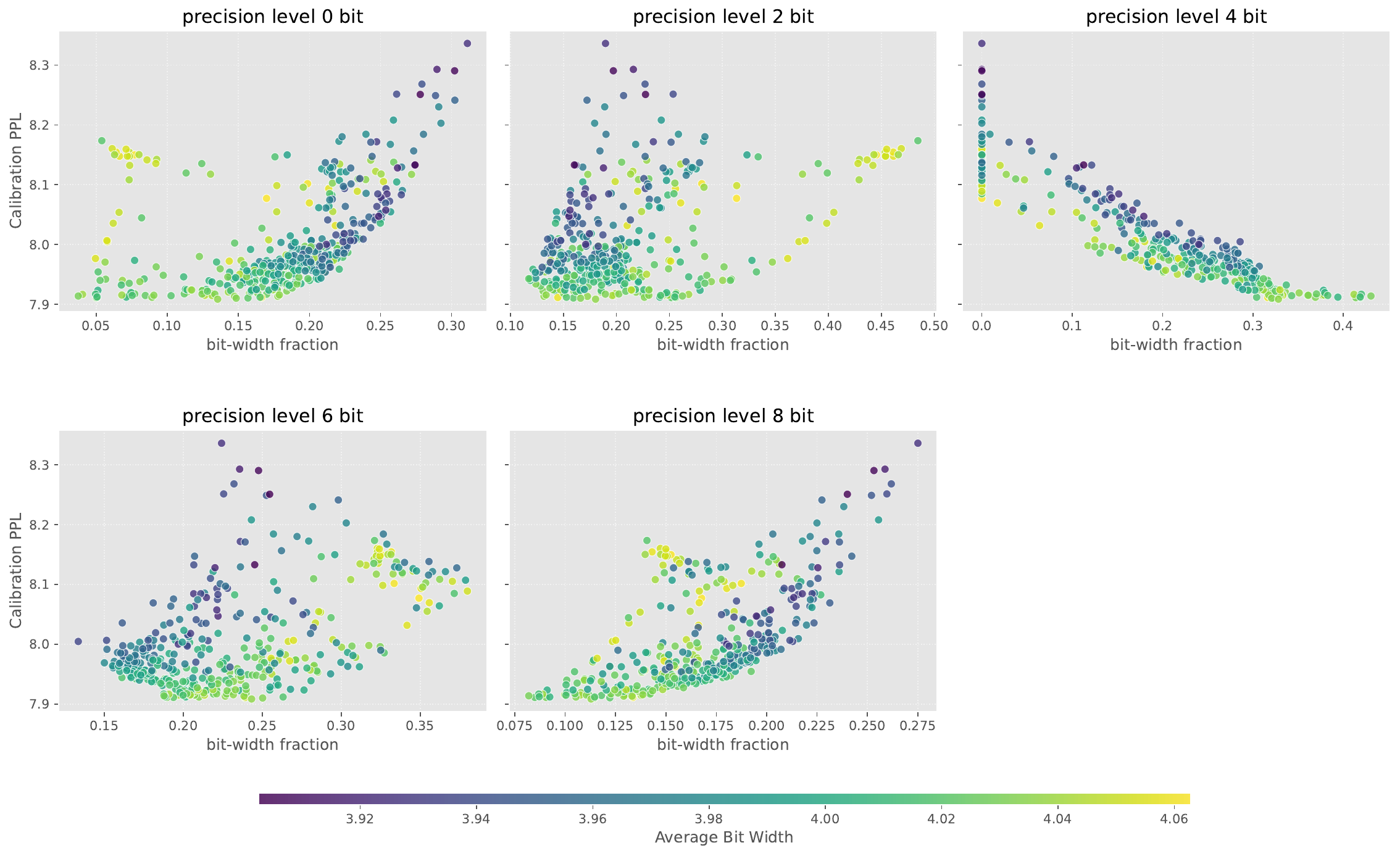}
    \caption{
    \textbf{Sensitivity of calibration perplexity to precision fractions $p_i$.   
    }
    Each panel illustrates the impact of a specific bit-width allocation on Llama3-1B performance, constrained to a target effective bit width $b^*=4.0 \pm 0.1$ ($b_\mathrm{max}=8$). We observe divergent scaling behaviors across the precision levels: boundary bit widths ($b_0$, $b_1$, and $b_4$) show a positive correlation with perplexity, suggesting that excessive allocation to extremes, degrades performance. In contrast, the 4-bit fraction ($b_2$) exhibits a strong inverse correlation. The 6-bit fraction ($b_3$) displays non-monotonic behavior, indicating a localized optimal range. These empirical trends provide the basis for the reduced hyperparameter search space. 
    }
    \label{fig:hyperparameterSweep}
\end{figure*}

A representative sweep for Llama3-1B at a 4-bit target (Fig.~\ref{fig:hyperparameterSweep}) suggests that certain sub-spaces are significantly more performant. For instance, high-performing configurations often fall within the bounds: $p_0 < 0.17$, $p_1 < 0.25$, $p_2>0.3$, $0.2<p_3<0.25$ and $p_4 < 0.13$. While these specific bounds shift with the target $b^*$, they provide a useful prior for narrowing the search space.

\paragraph{Recommended Configuration Workflow}
Based on these findings, we recommend the following four-step procedure to obtain a suitable hyperparameter instantiation:

\begin{enumerate}
    \item \textbf{Initialize:} Define the target $b^*$ and a small tolerance $\epsilon$ (e.g., 0.05).
    \item \textbf{Sample:} Use a constrained random sampler (e.g., based on a Dirichlet distribution) to generate $\sim 15$ candidate vectors $\mathrm{P}$ that satisfy the simplex constraints and align with the sparsity regimes identified above.
    \item \textbf{Calibrate:} Perform a single calibration forward pass for each $\mathrm{P}$ to determine the layer-wise thresholds and the resulting calibration PPL.
    \item \textbf{Select:} Deploy the configuration that yields the lowest calibration PPL.
\end{enumerate}

This empirical approach effectively identifies high-performance distributions without the need for an exhaustive search. Future work may further automate this process through the development of learnable classifiers or direct threshold optimization.

\newpage
\section{Appendix: Extended Comparative Analysis}

This section provides additional results comparing GRINQH with established state-of-the-art quantization methods (Section~\ref{app:extendedBenchmarks}) and existing multi-precision approaches (Section~\ref{app:comparisonAgainstMultiprecisionMethods}).

\subsection{Extended Accuracy Benchmarks} \label{app:extendedBenchmarks}

\subsubsection{Extended Pareto Frontier Plots} \label{app:paretoFrontier}

\begin{figure*}[ht]
    \centering    
    \includegraphics[width=\textwidth]{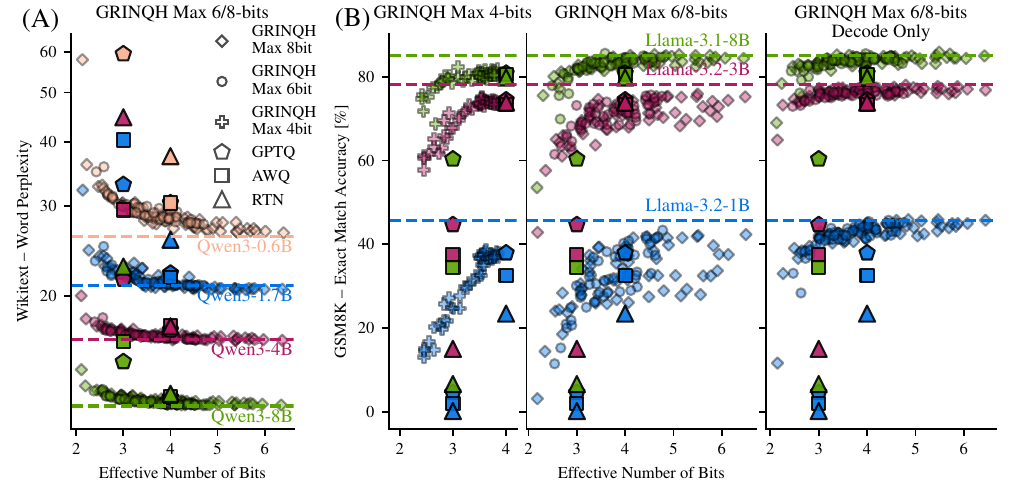}
    \caption{
    \textbf{GRINQH redefines the quantization Pareto frontier across model families and scales.} Dashed lines indicate BF16 baselines. GRINQH data points represent a sweep of precision distributions compared against iso-bit symmetric RTN, GPTQ, and AWQ baselines.
    \textbf{(A)} WikiText-2 perplexity vs. effective bit width for the Qwen3 family using $b_\text{max} \in \{6,8\}$. GRINQH outperforms SOTA counterparts, maintaining representational stability deep into the 2–3 bit regime.
    \textbf{(B)} GSM8K CoT accuracy vs. effective bit width for the Llama3 family. \textit{(Left panel)} GRINQH ($b_\text{max}=4$) sweep on GPTQ baseline.
    GRINQH matches the GPTQ 4-bit baseline and dominates specialized 3-bit RTN, AWQ, and GPTQ baselines. 
     \textit{(Middle panel)} GRINQH with $b_\text{max} \in \{6,8\}$ on RTN baseline. Our framework effectively recovers BF16 performance for 3B and 8B model scales in low-bit regime. 
     \textit{(Right panel)} Decoupled inference strategy ( $b_\text{max} \in \{6,8\}$ on RTN baseline) where dynamic precision loading is enabled exclusively for the memory-bound decoding stage.
     This real-case scenario boosts accuracy even further, particularly for smaller models, demonstrating the significant benefit of stage-aware optimization.   
    }
    \label{fig:pareto_joint_appendix}
\end{figure*}

Figure~\ref{fig:pareto_joint_appendix} presents the complementary Pareto frontier analysis with model families reversed: the Qwen3 family evaluated on WikiText-2 (A) and the Llama3 family on GSM8K (B). These results demonstrate that GRINQH consistently redefines the Pareto frontier across model families, scales, and benchmarks.

Notably, at an effective bit width of 3 bits, all GRINQH configurations (regardless of hyperparameter settings) outperform all static-precision SOTA counterparts on both WikiText-2 and GSM8K across all investigated peak memory settings ($b_\text{max} \in \{4,6,8\}$). For the decoding-heavy GSM8K CoT benchmark, where GRINQH’s mixed-precision logic is applied exclusively to the decoding stage (see Figure~\ref{fig:pareto_joint_appendix}B, right panel), BF16 baseline performance is effectively recovered even at 3 effective bits for the larger Llama3 models (3B and 8B). 

The resulting frontiers extend significantly further along the baseline and exhibit higher density compared to runs where mixed precision is applied to both inference phases (Figure~\ref{fig:pareto_joint_appendix}B, middle panel). This further underlines the benefits of stage-aware optimization, specifically treating the prefill and decoding stages as distinct computational regimes.

\subsubsection{Extended Table Model size comparison}

Tab.~\ref{tab:qwen-comparison-appendix} and Tab.~\ref{tab:llama-comparison-appendix} provide a comprehensive comparison across model scales ranging from 0.6B to 8B for the Qwen3 and Llama3 families at effective bit widths of 4, 3, and 2 bits. In the 2-bit regime, we report performance exclusively for GRINQH, as the perplexity of baseline SOTA methods diverges significantly at this level, and their performance on multiple-choice benchmarks falls to chance-level accuracy. Across all evaluated model sizes and families, GRINQH demonstrates superior performance over all baselines, consistently achieving lower perplexity on WikiText-2 and LAMBADA (LMB), as well as higher average accuracy across all evaluated bit widths.

\begin{table*}[ht]
\centering
\caption{Benchmark results for the Qwen3 model family.}
\resizebox{\textwidth}{!}{
\begin{tabular}{ l c l c c c c c c c c c } 
\toprule 
 & Eff. & Method  & WikiText2  & LMB  & LMB  & ARC-C  & HellaSwag  & BoolQ  & MMLU  & GSM8K  & Average \\ 
 & Bits &   & ppl $\downarrow$  & ppl $\downarrow$  & acc $\uparrow$  & acc $\uparrow$  & acc $\uparrow$  & acc $\uparrow$  & acc $\uparrow$  & acc $\uparrow$  & acc $\uparrow$ \\ 
\midrule 
\midrule 
\multirow{14}{*}{\rotatebox{90}{Qwen3-0.6B}} & 16 & BF16-Baseline              & 26.194  & 24.705  & 40.91  & 34.47  & 47.39  & 64.16  & 39.90  & 41.93  & 44.79  \\ 
\cmidrule{2-12}
 & 4 & GPTQ              & 30.424  & 37.620  & 36.35  & 30.29  & 44.89  & 61.93  & 33.74  & 29.80  & 39.50  \\ 
 & 4 & AWQ               & 30.404  & 42.145  & 33.92  & 29.78  & 44.23  & 50.40  & 37.43  & 26.91  & 37.11  \\ 
 & 4 & RTN               & 37.537  & 71.377  & 26.06  & 29.52  & 43.60  & 66.73  & 31.24  & 24.56  & 36.95  \\ 
 & 4.04 & GRINQH-8b-RTN     & 27.123  & 26.535  & 39.71  & 34.98  & 46.80  & 61.71  & 39.41  & 40.03  & 43.77  \\ 
 & 3.90 & GRINQH-6b-RTN     & 28.132  & 27.138  & 39.51  & 32.59  & 46.03  & 58.65  & 40.81  & 36.77  & 42.39  \\ 
\cmidrule{2-12}
 & 3 & GPTQ              & 59.460  & 431.130  & 14.52  & 26.96  & 37.33  & 52.51  & 24.28  & 1.52  & 26.19  \\ 
 & 3 & AWQ               & 94.550  & 893.252  & 8.21  & 24.06  & 34.23  & 59.57  & 23.53  & 0.83  & 25.07  \\ 
 & 3 & RTN               & 905.450  & 19216.238  & 2.19  & 22.78  & 29.37  & 42.32  & 24.28  & 0.00  & 20.16  \\ 
 & 3.01 & GRINQH-8b-RTN     & 29.879  & 32.596  & 37.88  & 32.51  & 45.37  & 67.03  & 38.99  & 32.60  & 42.40  \\ 
 & 2.94 & GRINQH-6b-RTN     & 29.995  & 31.322  & 37.01  & 31.57  & 44.73  & 55.44  & 36.38  & 36.01  & 40.19  \\ 
 & 3.03 & GRINQH-4b-RTN     & 39.235  & 76.691  & 25.42  & 29.86  & 42.86  & 64.04  & 30.17  & 22.44  & 35.80  \\ 
 & 3.06 & GRINQH-4b-GPTQ    & 31.575  & 40.674  & 35.55  & 28.50  & 44.34  & 61.10  & 32.52  & 26.46  & 38.08  \\ 
\midrule 
\midrule 
\multirow{14}{*}{\rotatebox{90}{Qwen3-1.7B}} & 16 & BF16-Baseline              & 20.974  & 12.182  & 51.47  & 43.00  & 60.41  & 77.49  & 55.59  & 73.24  & 60.20  \\ 
\cmidrule{2-12}
 & 4 & GPTQ              & 22.162  & 13.456  & 50.57  & 39.76  & 58.64  & 77.89  & 51.35  & 61.33  & 56.59  \\ 
 & 4 & AWQ               & 21.748  & 16.740  & 45.92  & 39.68  & 58.61  & 72.72  & 52.31  & 62.09  & 55.22  \\ 
 & 4 & RTN               & 25.778  & 23.280  & 39.90  & 39.85  & 57.91  & 76.09  & 52.05  & 54.97  & 53.46  \\ 
 & 4.05 & GRINQH-8b-RTN     & 20.942  & 12.773  & 50.94  & 43.52  & 60.17  & 77.13  & 55.08  & 71.87  & 59.78  \\ 
 & 3.96 & GRINQH-6b-RTN     & 21.021  & 12.420  & 51.87  & 42.32  & 59.44  & 77.37  & 55.61  & 68.99  & 59.27  \\ 
\cmidrule{2-12}
 & 3 & GPTQ              & 33.027  & 57.929  & 31.69  & 30.03  & 49.40  & 68.41  & 37.55  & 11.83  & 38.15  \\ 
 & 3 & AWQ               & 40.370  & 229.731  & 20.76  & 30.63  & 48.98  & 66.51  & 31.16  & 2.73  & 33.46  \\ 
 & 3 & RTN               & 88.595  & 1867.752  & 8.34  & 27.30  & 40.78  & 55.38  & 29.52  & 1.44  & 27.13  \\ 
 & 3.02 & GRINQH-8b-RTN     & 22.415  & 13.827  & 48.61  & 42.06  & 58.52  & 73.67  & 53.80  & 64.97  & 56.94  \\ 
 & 3.07 & GRINQH-6b-RTN     & 21.739  & 13.160  & 50.42  & 42.66  & 58.67  & 76.94  & 54.47  & 64.29  & 57.91  \\ 
 & 2.90 & GRINQH-4b-RTN     & 25.544  & 24.993  & 39.08  & 39.08  & 57.11  & 76.61  & 51.43  & 50.95  & 52.38  \\ 
 & 3.05 & GRINQH-4b-GPTQ    & 22.415  & 14.031  & 50.11  & 39.68  & 57.72  & 77.22  & 50.90  & 57.24  & 55.48  \\ 
\midrule 
\midrule 
\multirow{11}{*}{\rotatebox{90}{Qwen3-4B}} & 16 & BF16-Baseline              & 16.427  & 7.308  & 60.26  & 53.84  & 68.48  & 84.95  & 68.37  & 83.40  & 69.88  \\ 
\cmidrule{2-12}
 & 4 & GPTQ              & 17.169  & 8.090  & 58.16  & 53.16  & 67.60  & 83.58  & 66.66  & 79.15  & 68.05  \\ 
 & 4 & AWQ               & 17.507  & 9.941  & 55.97  & 51.45  & 66.63  & 83.76  & 65.67  & 79.38  & 67.14  \\ 
 & 4 & RTN               & 17.459  & 10.068  & 55.56  & 50.94  & 66.63  & 83.91  & 66.14  & 77.03  & 66.70  \\ 
 & 4.05 & GRINQH-8b-RTN     & 16.413  & 7.093  & 60.76  & 52.99  & 68.38  & 84.89  & 68.26  & 84.15  & 69.91  \\ 
 & 4.08 & GRINQH-6b-RTN     & 16.486  & 7.216  & 60.04  & 52.05  & 67.99  & 85.41  & 68.01  & 83.55  & 69.51  \\ 
\cmidrule{2-12}
 & 3 & GPTQ              & 21.560  & 16.553  & 44.42  & 41.30  & 60.52  & 80.92  & 56.40  & 46.17  & 54.95  \\ 
 & 3 & AWQ               & 29.480  & 129.466  & 24.26  & 40.19  & 56.84  & 74.31  & 50.97  & 22.67  & 44.87  \\ 
 & 3 & RTN               & 44.715  & 962.708  & 10.01  & 35.41  & 50.62  & 53.24  & 40.41  & 4.40  & 32.35  \\ 
 & 3.00 & GRINQH-8b-RTN     & 16.828  & 7.566  & 59.77  & 51.11  & 67.63  & 84.53  & 67.37  & 84.38  & 69.13  \\ 
 & 3.06 & GRINQH-6b-RTN     & 16.807  & 7.011  & 60.49  & 50.94  & 66.89  & 84.65  & 67.28  & 81.88  & 68.69  \\ 
 & 3.03 & GRINQH-4b-RTN     & 17.514  & 9.913  & 55.07  & 49.23  & 66.09  & 83.18  & 65.82  & 74.98  & 65.73  \\ 
 & 3.04 & GRINQH-4b-GPTQ    & 17.184  & 7.871  & 58.32  & 52.05  & 67.25  & 83.09  & 66.07  & 78.77  & 67.59  \\ 
\cmidrule{2-12}
 & 2.12 & GRINQH-8b-RTN     & 20.032  & 10.956  & 53.31  & 46.93  & 61.75  & 79.17  & 61.18  & 58.38  & 60.12  \\ 
 & 2.11 & GRINQH-4b-RTN     & 18.878  & 12.405  & 51.10  & 35.32  & 63.03  & 81.53  & 61.80  & 67.48  & 60.04  \\ 
 & 2.12 & GRINQH-4b-GPTQ    & 19.107  & 9.546  & 54.69  & 48.89  & 63.47  & 80.76  & 63.19  & 73.24  & 64.04  \\ 
\midrule 
\midrule 
\multirow{32}{*}{\rotatebox{90}{Qwen3-8B}} & 16 & BF16-Baseline              & 12.200  & 4.594  & 65.11  & 56.31  & 74.82  & 86.64  & 72.90  & 88.32  & 74.02  \\ 
\cmidrule{2-12}
  & 4 & GPTQ              & 12.680  & 5.169  & 63.13  & 54.35  & 74.42  & 86.79  & 71.58  & 82.79  & 72.18  \\ 
 & 4 & AWQ               & 12.702  & 5.091  & 63.32  & 54.61  & 73.48  & 85.78  & 71.00  & 82.41  & 71.77  \\ 
 & 4 & RTN               & 12.887  & 5.400  & 62.00  & 51.88  & 73.99  & 85.63  & 71.09  & 74.75  & 69.89  \\ 
 & 4 & QuaRot-GPTQ       & 12.581  & 4.810  & 64.54  & 56.48  & 73.76  & 85.87  & 71.00  & 86.28  & 72.99  \\ 
 & 4 & AutoRound         & 13.299  & 5.942  & 60.28  & 54.61  & 73.78  & 86.02  & 71.96  & 84.08  & 71.79  \\ 
 & 4 & NVFP4-G16         & 12.382  & 5.066  & 63.75  & 57.76  & 74.39  & 86.27  & 71.83  & 86.13  & 73.36  \\ 
 & 4 & AnyPrecisionLLM   & 12.820  & 4.695  & 64.74  & 55.80  & 73.75  & 86.70  & 71.51  & 87.64  & 73.36  \\ 
 & 4    & SliM-LLM    & 12.611     & 5.201     & 61.60 &  53.84 &  74.39     & 86.39  & 71.37 & 87.41  & 72.50 \\
 & 3.97 & GRINQH-8b-RTN     & 12.291  & 4.581  & 65.17  & 55.63  & 74.60  & 86.48  & 72.75  & 87.57  & 73.70  \\ 
 & 3.95 & GRINQH-8b-AnyPrec  & 12.234  & 4.439  & 65.11  & 56.40  & 74.77  & 86.67  & 72.82  & 87.72  & 73.91  \\ 
 & 3.94 & GRINQH-6b-RTN     & 12.265  & 4.580  & 65.09  & 56.14  & 74.74  & 86.85  & 72.85  & 87.19  & 73.81  \\ 
\cmidrule{2-12}
 & 3 & GPTQ              & 14.886  & 6.957  & 56.26  & 47.01  & 68.87  & 83.76  & 64.31  & 67.78  & 64.67  \\ 
 & 3 & AWQ               & 16.327  & 14.239  & 43.68  & 43.34  & 66.05  & 80.67  & 61.70  & 59.67  & 59.19  \\ 
 & 3 & RTN               & 22.844  & 41.288  & 28.26  & 38.82  & 61.39  & 70.95  & 53.79  & 15.54  & 44.79  \\ 
 & 3 & QuaRot-GPTQ       & 14.282  & 7.297  & 55.13  & 48.81  & 69.03  & 83.82  & 64.53  & 68.84  & 65.03  \\ 
 & 3 & AutoRound         & 15.079  & 7.307  & 54.49  & 49.57  & 69.99  & 83.91  & 66.96  & 67.78  & 65.45  \\ 
 & 3 & AnyPrecisionLLM   & 15.231  & 6.499  & 59.62  & 52.56  & 69.96  & 86.09  & 68.15  & 79.30  & 69.28  \\ 
 & 3    & SliM-LLM    & 14.884     & 5.582     & 60.35 & 46.93  &  70.45     & 84.19  & 66.27 & 77.63  & 67.64 \\
 & 3.07 & GRINQH-8b-RTN     & 12.610  & 4.512  & 65.42  & 55.38  & 73.63  & 85.87  & 71.66  & 88.02  & 73.33  \\ 
 & 3.06 & GRINQH-6b-RTN     & 12.581  & 4.527  & 65.65  & 55.63  & 73.65  & 86.61  & 71.71  & 85.60  & 73.14  \\ 
 & 3.03 & GRINQH-4b-RTN     & 13.020  & 5.304  & 62.41  & 53.50  & 73.81  & 85.75  & 70.80  & 79.38  & 70.94  \\ 
 & 3.03 & GRINQH-4b-GPTQ    & 12.841  & 5.101  & 63.21  & 53.16  & 73.90  & 86.57  & 71.12  & 82.56  & 71.75  \\ 
 & 3.06 & GRINQH-8b-AnyPrec  & 12.403  & 4.348  & 65.48  & 56.14  & 74.06  & 86.76  & 72.61  & 87.41  & 73.74  \\ 
\cmidrule{2-12}
 & 2 & AnyPrecisionLLM   & 109.867  & 669.490  & 14.24  & 34.22  & 47.28  & 67.86  & 38.61  & 0.61  & 33.80  \\ 
 & 2    & SliM-LLM    & 100.974    & 212.912   & 25.05 & 26.19  &  40.83     & 63.88  & 30.57 &  2.12  & 31.44 \\
 & 2.25 & GRINQH-8b-RTN     & 13.356  & 5.117  & 63.50  & 53.41  & 72.35  & 86.21  & 70.11  & 84.00  & 71.60  \\ 
 & 2.22 & GRINQH-8b-AnyPrec  & 12.947  & 4.209  & 66.14  & 56.14  & 72.53  & 86.70  & 71.10  & 87.04  & 73.27  \\ 
 & 2.11 & GRINQH-4b-RTN     & 14.365  & 5.885  & 61.11  & 49.74  & 70.39  & 84.53  & 67.04  & 75.82  & 68.11  \\ 
 & 2.11 & GRINQH-4b-GPTQ    & 14.086  & 5.969  & 61.77  & 52.22  & 70.53  & 85.20  & 68.60  & 77.86  & 69.36  \\ 
 \midrule 
\bottomrule 
\end{tabular} 
}\label{tab:qwen-comparison-appendix}
\end{table*}

\begin{table*}[ht]
\centering
\caption{Benchmark results for the Llama3 model family.}
\resizebox{\textwidth}{!}{
\begin{tabular}{ l c l c c c c c c c c c } 
\toprule 
 & Eff. & Method  & WikiText2  & LMB  & LMB  & ARC-C  & HellaSwag  & BoolQ  & MMLU  & GSM8K  & Average \\ 
 & Bits &   & ppl $\downarrow$  & ppl $\downarrow$  & acc $\uparrow$  & acc $\uparrow$  & acc $\uparrow$  & acc $\uparrow$  & acc $\uparrow$  & acc $\uparrow$  & acc $\uparrow$ \\ 
\midrule 
\midrule 
\multirow{14}{*}{\rotatebox{90}{Llama-3.2-1B-Instruct}} & 16 & BF16-Baseline              & 15.813  & 6.580  & 61.07  & 37.71  & 60.85  & 69.20  & 45.95  & 45.72  & 53.42  \\ 
\cmidrule{2-12}
 & 4 & GPTQ              & 18.685  & 8.854  & 55.48  & 37.88  & 59.40  & 69.02  & 45.16  & 37.91  & 50.81  \\ 
 & 4 & AWQ               & 19.337  & 8.499  & 56.05  & 37.63  & 58.98  & 66.57  & 44.74  & 32.60  & 49.43  \\ 
 & 4 & RTN               & 18.958  & 10.183  & 53.21  & 34.98  & 56.39  & 66.39  & 41.79  & 23.43  & 46.03  \\ 
 & 4.06 & GRINQH-8b-RTN     & 16.333  & 6.744  & 60.10  & 36.35  & 59.94  & 69.60  & 46.40  & 41.55  & 52.32  \\ 
 & 3.94 & GRINQH-6b-RTN     & 16.443  & 6.932  & 59.89  & 37.80  & 59.24  & 69.27  & 45.54  & 37.23  & 51.49  \\ 
\cmidrule{2-12}
 & 3 & GPTQ              & 32.059  & 33.060  & 35.51  & 31.74  & 52.08  & 63.79  & 35.84  & 4.55  & 37.25  \\ 
 & 3 & AWQ               & 62.497  & 61.082  & 27.25  & 30.38  & 49.28  & 56.15  & 29.13  & 1.97  & 32.36  \\ 
 & 3 & RTN               & 61.012  & 393.156  & 12.61  & 26.62  & 39.28  & 60.80  & 25.95  & 0.15  & 27.57  \\ 
 & 3.04 & GRINQH-8b-RTN     & 17.372  & 7.852  & 56.86  & 37.03  & 58.26  & 68.38  & 44.38  & 34.12  & 49.84  \\ 
 & 2.97 & GRINQH-6b-RTN     & 17.766  & 8.923  & 54.65  & 35.92  & 56.79  & 68.04  & 42.94  & 28.81  & 47.86  \\ 
 & 3.04 & GRINQH-4b-RTN     & 19.430  & 10.797  & 52.42  & 34.56  & 55.66  & 65.84  & 40.99  & 21.76  & 45.20  \\ 
 & 2.95 & GRINQH-4b-GPTQ    & 18.641  & 11.009  & 52.14  & 36.60  & 55.76  & 66.88  & 41.50  & 24.64  & 46.25  \\ 
\cmidrule{2-12}
 & 2.12 & GRINQH-8b-RTN     & 33.031  & 41.759  & 31.54  & 31.23  & 49.16  & 62.45  & 29.97  & 3.11  & 34.57  \\ 
\midrule 
\midrule 
\multirow{14}{*}{\rotatebox{90}{Llama-3.2-3B-Instruct}} & 16 & BF16-Baseline              & 12.276  & 4.821  & 67.18  & 46.25  & 70.44  & 78.41  & 60.29  & 78.24  & 66.80  \\ 
\cmidrule{2-12}
 & 4 & GPTQ              & 127.864  & 5.262  & 65.11  & 44.62  & 70.51  & 78.10  & 60.11  & 74.45  & 65.49  \\ 
 & 4 & AWQ               & 13.257  & 5.505  & 63.54  & 45.31  & 70.61  & 78.59  & 60.25  & 73.39  & 65.28  \\ 
 & 4 & RTN               & 13.246  & 5.570  & 63.71  & 44.54  & 70.50  & 77.55  & 59.31  & 73.84  & 64.91  \\ 
 & 3.92 & GRINQH-8b-RTN     & 12.767  & 4.746  & 67.05  & 44.71  & 71.11  & 76.06  & 59.96  & 72.40  & 65.22  \\ 
 & 3.89 & GRINQH-6b-RTN     & 12.692  & 4.753  & 67.24  & 44.62  & 71.29  & 77.49  & 60.60  & 72.93  & 65.70  \\ 
\cmidrule{2-12}
 & 3 & GPTQ              & 6190.919  & 9.138  & 53.83  & 39.42  & 65.61  & 68.01  & 51.18  & 44.66  & 53.79  \\ 
 & 3 & AWQ               & 24.194  & 11.161  & 50.55  & 40.61  & 64.64  & 73.58  & 44.92  & 37.45  & 51.96  \\ 
 & 3 & RTN               & 23.948  & 15.715  & 46.03  & 35.49  & 61.19  & 66.02  & 41.15  & 15.01  & 44.15  \\ 
 & 3.00 & GRINQH-8b-RTN     & 13.077  & 4.915  & 66.37  & 43.60  & 70.51  & 75.23  & 58.99  & 69.52  & 64.04  \\ 
 & 2.93 & GRINQH-6b-RTN     & 13.256  & 5.110  & 65.98  & 43.17  & 70.30  & 75.60  & 58.32  & 68.46  & 63.64  \\ 
 & 2.89 & GRINQH-4b-RTN     & 13.692  & 5.510  & 64.66  & 43.94  & 69.79  & 74.56  & 57.27  & 68.69  & 63.15  \\ 
 & 2.88 & GRINQH-4b-GPTQ    & 13.909  & 5.264  & 65.88  & 40.36  & 69.90  & 76.94  & 57.20  & 67.70  & 63.00  \\ 
\cmidrule{2-12}
 & 2.12 & GRINQH-8b-RTN     & 18.303  & 8.427  & 57.56  & 38.05  & 65.67  & 63.88  & 49.32  & 42.84  & 52.89  \\ 
\midrule 
\midrule 
\multirow{28}{*}{\rotatebox{90}{Llama-3.1-8B-Instruct}} & 16 & BF16-Baseline              & 8.642  & 3.404  & 74.19  & 55.20  & 79.26  & 84.19  & 68.12  & 85.14  & 74.35  \\ 
\cmidrule{2-12} 
 & 4 & GPTQ              & 9.409  & 3.644  & 72.85  & 54.52  & 78.59  & 84.19  & 66.16  & 80.82  & 72.85  \\ 
 & 4 & AWQ               & 9.525  & 3.911  & 70.02  & 51.37  & 78.63  & 84.59  & 66.69  & 80.59  & 71.98  \\ 
 & 4 & RTN               & 9.665  & 3.903  & 70.11  & 50.77  & 78.62  & 84.65  & 64.65  & 79.83  & 71.44  \\ 
 & 4 & QuaRot-GPTQ       & 9.377  & 3.554  & 72.73  & 52.90  & 78.39  & 85.20  & 66.04  & 81.88  & 72.86  \\ 
 & 4 & AutoRound         & 9.615  & 3.894  & 70.46  & 54.18  & 78.42  & 84.28  & 65.28  & 80.89  & 72.25  \\ 
 & 4 & NVFP4-G16         & 9.272  & 3.408  & 73.86  & 53.84  & 78.75  & 84.13  & 66.36  & 82.49  & 73.24  \\ 
 & 4 & AnyPrecisionLLM   & 9.444  & 3.714  & 69.63  & 55.20  & 78.49  & 85.14  & 66.69  & 81.88  & 72.84  \\ 
 & 4    & SliM-LLM   & 9.354      & 3.463     & 72.19 &  52.99 &  78.45     & 84.56  & 66.07 & 82.64  & 72.82 \\
 & 4.01 & GRINQH-8b-RTN     & 8.971  & 3.377  & 73.78  & 54.95  & 79.27  & 85.02  & 67.88  & 85.14  & 74.34  \\ 
 & 3.89 & GRINQH-8b-AnyPrec  & 8.945  & 3.358  & 73.18  & 54.86  & 79.18  & 85.29  & 68.07  & 84.53  & 74.19  \\ 
 & 3.92 & GRINQH-6b-RTN     & 9.020  & 3.409  & 73.26  & 54.18  & 78.87  & 85.41  & 67.43  & 84.31  & 73.91  \\ 
 \cmidrule{2-12}
 & 3 & GPTQ              & 12.461  & 5.857  & 60.88  & 44.28  & 74.08  & 81.83  & 58.18  & 60.35  & 63.27  \\ 
 & 3 & AWQ               & 14.363  & 8.039  & 54.82  & 45.39  & 72.46  & 78.62  & 50.52  & 34.50  & 56.05  \\ 
 & 3 & RTN               & 19.133  & 7.427  & 60.10  & 46.25  & 68.09  & 73.76  & 49.05  & 6.60  & 50.64  \\ 
 & 3 & QuaRot-GPTQ       & 12.798  & 5.768  & 63.07  & 47.70  & 73.24  & 80.70  & 58.13  & 57.54  & 63.40  \\ 
 & 3 & AutoRound         & 11.812  & 5.239  & 64.43  & 49.15  & 75.61  & 81.41  & 57.83  & 49.58  & 63.00  \\ 
 & 3 & AnyPrecisionLLM   & 12.217  & 4.832  & 63.19  & 50.60  & 75.32  & 82.72  & 59.52  & 56.18  & 64.59  \\ 
 & 3    & SliM-LLM       & 12.010  & 4.853     & 65.53 & 50.00  &  73.65     & 82.35  & 58.09 & 68.46  & 66.35  \\
 & 2.98 & GRINQH-8b-RTN     & 9.304  & 3.485  & 73.03  & 54.44  & 78.50  & 85.20  & 66.50  & 81.50  & 73.19  \\ 
 & 2.96 & GRINQH-8b-AnyPrec  & 9.220  & 3.368  & 72.52  & 55.12  & 78.41  & 84.86  & 67.02  & 84.23  & 73.69  \\ 
 & 3.02 & GRINQH-6b-RTN     & 9.435  & 3.402  & 73.01  & 53.24  & 77.88  & 84.31  & 65.62  & 82.41  & 72.75  \\ 
 & 3.02 & GRINQH-4b-RTN     & 9.802  & 3.799  & 70.68  & 51.02  & 77.88  & 84.77  & 64.02  & 79.68  & 71.34  \\ 
 & 3.04 & GRINQH-4b-GPTQ    & 9.593  & 3.563  & 73.06  & 53.50  & 78.24  & 84.13  & 65.23  & 79.91  & 72.34  \\ 
 \cmidrule{2-12}
 & 2    & SliM-LLM   & 535.131    & 34205.477 & 1.43  & 24.74  &  27.52     & 37.86  & 23.01 &  2.96  & 19.59 \\
 & 2 & AnyPrecisionLLM   & 1953.566  & 11649.584  & 2.87  & 27.47  & 34.69  & 41.93  & 24.86  & 1.52  & 22.22  \\ 
 & 2.11 & GRINQH-8b-RTN     & 12.871  & 5.937  & 61.46  & 45.39  & 73.94  & 78.56  & 55.36  & 53.53  & 61.37  \\ 
 & 2.01 & GRINQH-8b-AnyPrec  & 10.465  & 3.717  & 69.24  & 52.30  & 75.40  & 83.98  & 61.76  & 71.65  & 69.06  \\ 
\midrule 
\bottomrule 
\end{tabular} 

} \label{tab:llama-comparison-appendix}
\end{table*}

\subsection{Comparisons to Multi-Precision Methods} \label{app:comparisonAgainstMultiprecisionMethods}

Existing any-precision and mixed-precision frameworks generally target three distinct paradigms: (i)~improved nested or multi-precision weight representations (e.g., any-precision LLM~\cite{AnyPrecisionLLM24}, MatQuant~\cite{Nair25}, MatGPTQ~\cite{MatGPTQ26}), (ii)~static mixed-precision allocation at the layer or block level (e.g., Mix’n’Match~\cite{MatFormer23, Nair25, MatGPTQ26}, SliM-LLM~\cite{SliM-LLM24}), or (iii)~coarse-grained dynamic scheduling (e.g., PMPD~\cite{Chen24PMPD}, where precision is token-dependent). GRINQH is complementary to paradigm (i), as its quantization-agnostic framework can directly benefit from more sophisticated weight-quantization methods (see Section~\ref{app:evaluationOfNestedWeightFormats} for more details). Moreover, in contrast to (ii) and (iii), GRINQH performs activation-dependent, input-channel-wise precision allocation within each decoding step. This finer granularity is a key advantage, as it enables the system to adapt dynamically to token-specific outlier shifts within each vector-matrix multiplication (VMM), rather than relying on static saliency or coarse-grained decisions.

Some methods, however, combine aspects of (i) and (ii). We therefore divide the methods into three categories for a more detailed comparison:

\paragraph{Representation-Focused and Complementary Methods.}
Any-precision LLM focuses on improving nested multi-precision weight construction respectively by reformulating QAT for improved truncation robustness and iterative K-means based incremental upscaling. However, both methods do not provide a runtime mixed-precision scheduling logic. Consequently, these methods are complementary to GRINQH; while they improve the underlying base weights, GRINQH provides the dynamic allocation and efficient loading mechanisms required for inference. Tab.~\ref{tab:8B_comparison} and Section~\ref{app:evaluationOfNestedWeightFormats} highlight the complementary nature by showing the performance increase of GRINQH when using any-precision LLM weights compared to uniform quantization techniques such as RTN and GPTQ.

\paragraph{Nested Methods with Static Coarse-Mixed Precision Allocation.}
MatQuant~\cite{Nair25} and MatGPTQ~\cite{MatGPTQ26} optimize the base nested representation and employ Mix’n’Match-style \cite{MatFormer23} layer-wise static bit allocation. GRINQH remains complementary to these approaches by instead optimizing runtime channel-wise allocation (see Section~\ref{app:evaluationOfNestedWeightFormats} for GRINQH based on MatGPTQ weights). Notably, MatQuant-style mixed-precision inference does not outperform optimized fixed-precision baselines in \cite{Nair25} (Fig. 1(b) and 2)), whereas GRINQH achieves significant gains over fixed-precision baselines (see Tab.~\ref{tab:8B_comparison}). Tab.~\ref{tab:llama-comparison-appendix-matgptq} shows that GRINQH outperforms both vanilla MatGPTQ and its optimized mixed-precision variants (MatGPTQ-EP-Mix’n’Match) across the full 2--4 bit range, with the gap widening significantly at 3 bits and below. For Llama-3.1-8B-Instruct at an effective bit width of 4 bits, GRINQH ($73.79\%$) reduces accuracy degradation relative to the BF16 baseline ($74.00\%$) by $6.3\times$ compared to MatGPTQ (72.67\%). At an effective bit width of 3 bits, this reduction in degradation is $16.9\times$ ($73.68\%$ vs. $68.58\%$).

\begin{figure*}[ht]
    \centering    
    \includegraphics[width=0.6\textwidth]{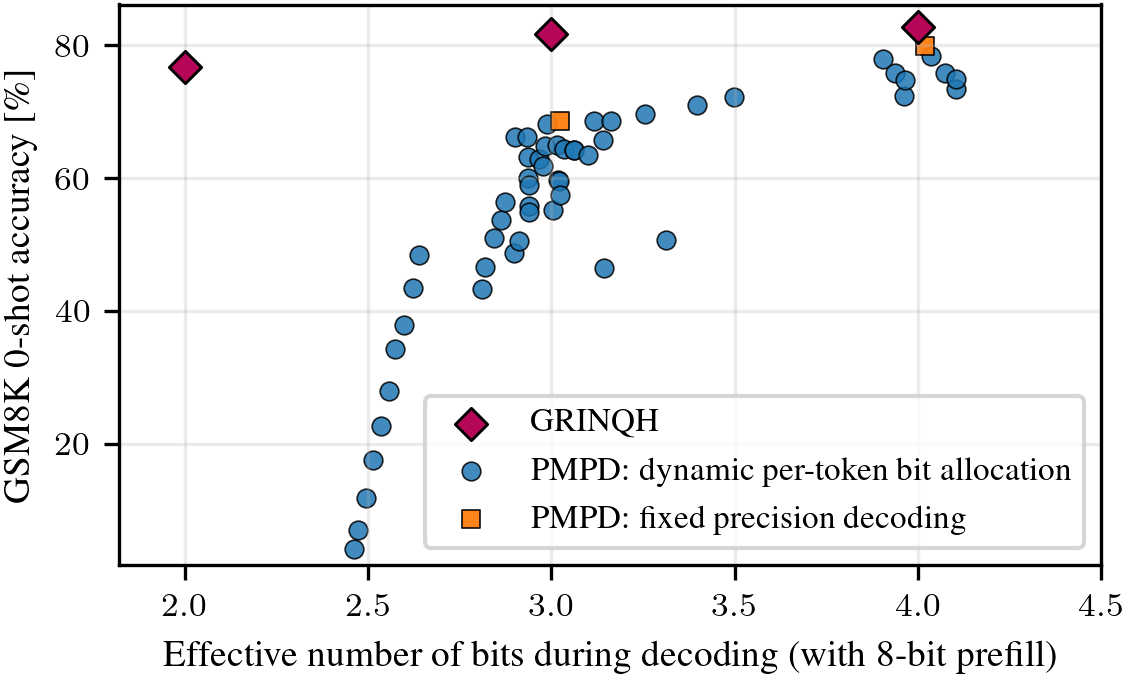}
    \caption{
        Comparison between GRINQH's fine-grained dynamic input-channel-wise bit allocation and PMPD's per-token bit allocation (decreasing precision at later token positions), as well as PMPD with static fixed-precision bit allocation during decoding. PMPD datapoints originate from a sweep over various configurations of 8/6/4/3/2-bit allocations during decoding and were generated using the original PMPD codebase. The task is GSM8K 0-shot. The model is Llama3.1-8B-Instruct. All datapoints use 8-bit prefill.
    }
    \label{fig:pmpd_comparison}
\end{figure*}

\paragraph{Alternative Mixed-Precision Scheduling.} 
SliM-LLM~\cite{SliM-LLM24} utilizes static per-block precision allocation based on saliency, which precludes adaptation to runtime outlier shifts during generation. Tab.~\ref{tab:8B_comparison} shows that GRINQH substantially outperforms SliM-LLM, especially in the 3-bit (73.69\% vs. 66.35\% on Llama-3.1-8B-Instruct) and 2-bit (69.06\% vs. 19.59\% on Llama-3.1-8B-Instruct) regime. This confirms the advantage of fine-grained dynamic precision allocation during runtime. PMPD~\cite{Chen24PMPD} is the most closely related prior work but utilizes coarser per-token dynamic scheduling. GRINQH’s input-channel-wise dynamic bit allocation is significantly more fine-grained and achieves substantially stronger results (e.g., 81.7\% vs. 68.7\% on GSM8K-0shot at 3 effective bits for Llama-3.1-8B-Instruct). We provide a detailed breakdown of these comparisons in Fig.~\ref{fig:pmpd_comparison}.

\begin{table*}[ht]
\centering
\caption{Benchmark results comparing GRINQH and MatGPTQ on the Llama3.1-8B-Instruct and Qwen3-8B models. For MatGPTQ, results are presented for both the same precision across all layers (MatGPTQ) and the optimized mixed-precision Mix'n'Match version (MatGPTQ-EP-Mix'n'Match), where layers operate at varying precision. The MatGPTQ results are taken from the original paper (Tabs. 6, 9, 19 and 20). Notably, 4-bit results for Qwen3-8B are not included, as Mix'n'Match results are not reported in the original paper. Across both models and all bit widths, GRINQH's dynamic input-channel-wise bit allocation consistently achieves superior average accuracy compared to both MatGPTQ's static fixed- and mixed-precision variants.}\label{tab:llama-comparison-appendix-matgptq}
\resizebox{\textwidth}{!}{
\begin{tabular}{ l c l c c c c c c} 
\toprule 
 & Eff. & Method & ARC-C & ARC-E & HellaSwag & PIQA~\cite{bisk2020piqa}  & Winogrande~\cite{sakaguchi2021winogrande} & Average \\ 
 & Bits &   & acc $\uparrow$  & acc $\uparrow$  & acc $\uparrow$  & acc $\uparrow$  & acc $\uparrow$  & acc $\uparrow$ \\ 
\midrule 
\multirow{12}{*}{\rotatebox{90}{Llama-3.1-8B-Instruct}} & 16 & BF16-Baseline & 55.89 & 79.80 & 79.52 & 81.34 & 73.48 & 74.00 \\ 
\cmidrule{2-9}
 & 4.00 & MatGPTQ                & 52.82 & 79.84 & 77.40 & 80.96 & 72.06 & 72.62 \\
 & 4.00 & MatGPTQ-EP-Mix'n'Match & 53.84 & 78.58 & 77.60 & 79.76 & 73.56 & 72.67 \\
 & 3.96 & GRINQH-8b-RTN & 55.80 & 79.29 & 79.21 & 80.30 & 74.35 & 73.79 \\
 & 3.881 & GRINQH-6b-RTN & 52.90 & 81.14 & 78.95 & 80.25 & 74.66 & 73.58 \\
\cmidrule{2-9}
 & 3.00 & MatGPTQ & 46.25 & 70.92 & 72.92 & 76.44 & 71.27 & 67.56 \\
 & 3.00 & MatGPTQ-EP-Mix'n'Match & 47.53 & 71.46 & 73.53 & 77.86 & 72.53 & 68.58 \\
 & 2.95 & GRINQH-8b-RTN & 55.03 & 79.67 & 78.68 & 80.25 & 73.01 & 73.68 \\
 & 2.98 & GRINQH-6b-RTN & 53.58 & 78.49 & 77.75 & 79.54 & 73.16 & 72.50 \\
 & 3.01 & GRINQH-4b-GPTQ & 54.27 & 78.45 & 78.36 & 80.47 & 72.53 & 72.82 \\
\cmidrule{2-9}
  & 2.08 & GRINQH-8b-RTN & 46.08 & 72.18 & 73.78 & 78.02 & 67.40 & 67.49 \\
\midrule 
\midrule 
\multirow{8}{*}{\rotatebox{90}{Qwen3-8B}} & 16 & BF16-Baseline & 56.57 & 80.93 & 74.94 & 77.69 & 67.56 & 71.54 \\ 
\cmidrule{2-9}
 & 3.00 & MatGPTQ & 48.72 & 72.31 & 67.18 & 74.32 & 64.64 & 65.43 \\
 & 3.00 & MatGPTQ-EP-Mix'n'Match & 47.53 & 68.69 & 68.40 & 74.27 & 64.48 & 64.67 \\
 & 3.06 & GRINQH-8b-RTN & 55.12 & 78.66 & 73.50 & 76.71 & 65.90 & 69.98 \\
 & 3.05 & GRINQH-6b-RTN & 56.14 & 79.50 & 73.63 & 77.64 & 67.64 & 70.91 \\
 & 3.02 & GRINQH-4b-GPTQ & 55.20 & 78.20 & 73.96 & 77.37 & 66.61 & 70.27 \\
 \cmidrule{2-9}
 & 2.24 & GRINQH-8b-RTN & 52.99 & 78.45 & 72.21 & 77.31 & 67.72 & 69.74 \\
 & 2.09 & GRINQH-4b-GPTQ & 51.71 & 76.94 & 70.94 & 76.77 & 66.93 & 68.66 \\
\midrule 
\bottomrule 
\end{tabular} 
} 
\end{table*}

\newpage
\section{Appendix: Kernel Implementation and Profiling}
This section details the implementation of our Triton GEMV kernel, focusing on how it leverages the bit-planar layouts for sparse loading and bit-packing (Section~\ref{app:kernelImplementationDetails}). We then validate these design choices via roofline analysis (Section~\ref{app:rooflineAnalysis}). Furthermore, we provide a version of Table~\ref{tab:vllm_prefill} that includes standard deviations for all measurements: Tab.~\ref{tab:vllm_prefill_EXT}.

\subsection{Implementation Details} \label{app:kernelImplementationDetails}

\textbf{Sparse bit-loading:} To optimize memory access for vector-matrix multiplication, we store the original weight matrix in a transposed (reduction-dimension major) layout. This ensures that for any element of the input vector $x$, the corresponding weights for all output channels are stored row-contiguously. The output dimension $N_{out}$ is partitioned into $\nicefrac{b_\text{max}}{2}$ precision zones, each storing a specific 2-bit significance slice of the quantized weights. Each zone spans $\nicefrac{2 N_{out}}{b_\text{max}}$ columns and stores all 2-bit segments of the weight values for a given precision level. Loading weights for higher precision, therefore, requires incremental loads offset from each zone, allowing for sparse loads. Zone sizing is static, since a $b_\text{max}$ below 8-bit simply stops iterating at $\nicefrac{b_\text{max}}{2}$.

\textbf{Bit-packing:} We pack bit-segments of the same significance and of neighbouring weights within an input channel into uint32 format to maximize register occupancy and memory bandwidth utilization. This wrapper holds four 8-bit bytes, with each byte containing four spatially grouped 2-bit chunks. At runtime, this layout allows the GEMV kernel to load the equivalent of 16 partial weights of a single significance in a single 32-bit instruction. It also cuts the total amount of memory requests required by a factor of four.

\textbf{Grouping:} To maintain writing locality, we pre-group weights into a 16-block layout. This mechanic aligns the internal structure of the uint32 packets with their logical destination in the output vector $y$. By offloading this structural realignment to a pre-processing operation, the GEMV kernel can utilize bitwise OR and SHIFT operations to reconstruct weights, without any runtime data-movement overhead.

\subsection{Roofline Analysis} \label{app:rooflineAnalysis}
\begin{figure*}[ht]
    \centering    
    \includegraphics[width=\textwidth]{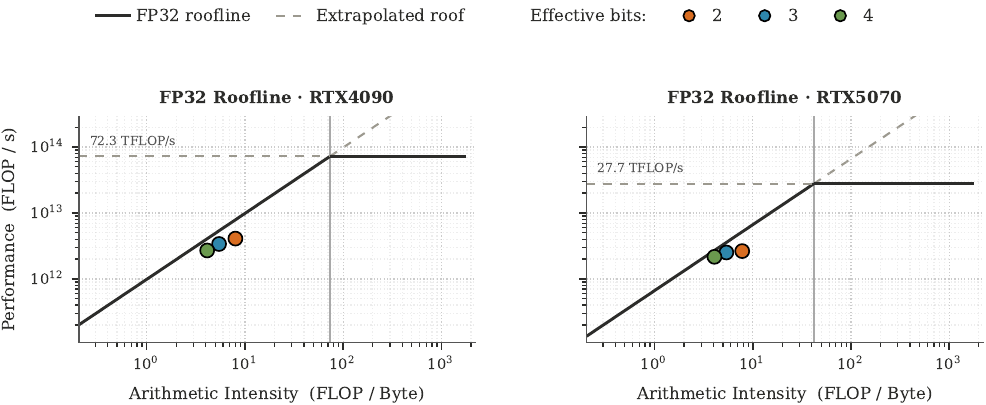}
    \caption{
        FP32 throughput over DRAM bytes loaded for different effective bit widths on consumer-grade devices, RTX4090 and RTX5070. The kernel was profiled with NVIDIA NSight Compute\cite{nvidia_nsight_compute_cli} in isolation using a standalone weight matrix of $W \in \mathbb{R}^{16{,}384 \times 4{,}096}$, a $b_\text{max}=8$ configuration for a target bit width of $4$ and uniformly distributed thresholds. Identical kernel hyperparameters were used, matching those of our end-to-end result runs for the same target bit width. The maximum throughput values are affected by changing clock rates during execution, but are close to the base rate for both GPUs ($73.24$ TFLOP/s and $27.46$ TFLOP/s respectively). All profiled kernels operate in the memory-bound regime. 
    }
    \label{fig:roofline}
\end{figure*}

Figure~\ref{fig:roofline} utilizes a roofline analysis to visualize the memory-bound regime of our kernel. Reducing the effective bit width decreases memory transfers, lowering total runtime and increasing both arithmetic intensity and compute throughput. Because the analysis tracks only floating-point operations, it does not capture integer math such as shifts and masking. This static overhead weighs on the effective 2-bit configuration, imposing a soft cap on latency-hidden compute utilization.

\begin{table}[h]
     \centering
     \caption{Tok/s performance over different effective bit widths and in:output ratios (prefill: (3000:1), decoding: (1:3000)) of tokens on random data. Relative performance is calculated against GPTQ 4-bit MARLIN.}\label{tab:vllm_prefill_EXT}
 \resizebox{\textwidth}{!}{
     \begin{tabular}{lcccc}
     \toprule
     \textbf{Kernel} & \textbf{Prefill Abs.} & \textbf{Prefill Rel.} & \textbf{Decoding Abs.} & \textbf{Decoding Rel.} \\
     \midrule
     Eff. 2-bit GRINQH (Max 8-bit) & \multirow{3}{*}{8186.78 $\pm$ 17.92} & \multirow{3}{*}{0.80x} & 204.64 $\pm$ 0.74 & 1.28x \\
     Eff. 3-bit GRINQH (Max 8-bit) & & & 182.18 $\pm$ 0.69 & 1.14x \\
     Eff. 4-bit GRINQH (Max 8-bit) & & & 157.56 $\pm$ 0.06 & 0.98x \\
     \midrule
     Eff. 4-bit GRINQH (Max 6-bit) & 8241.01 $\pm$ 28.23 & 0.81x & 165.10 $\pm$ 0.13 & 1.03x \\
     Eff. 3-bit GRINQH (Max 4-bit) & 8241.01 $\pm$ 28.23 & 0.81x & 165.10 $\pm$ 0.13 & 1.03x \\
    
     \midrule
     GPTQ 4-bit Marlin & 10170.38 $\pm$ 26.47 & 1.00x & 160.36 $\pm$ 0.19 & 1.00x \\
     RTN 8-bit Marlin & 8927.07 $\pm$ 23.35 & 0.88x & 100.21 $\pm$ 0.18 & 0.62x \\
     \bottomrule
     \end{tabular}
 }
\end{table}

\newpage
\section{Appendix: Calibration and Data Sensitivity} \label{app:calibration_analysis}

This section provides a detailed analysis of calibration resource requirements (Section~\ref{app:resourceConsumptionCalibration}), the sensitivity of the method to calibration data composition (Section~\ref{app:RepresentativenessCalibration}), and the fidelity of the calibration mechanism across downstream tasks (Section~\ref{app:calibrationFidelity}).
 
To ensure a controlled and fair comparison with SOTA baselines, we use a calibration setup (128 samples, 2048 context length) that matches the standard configuration of the methods we compare against, rather than optimizing for the best samples for GRINQH.

\subsection{Resource Consumption during Calibration} \label{app:resourceConsumptionCalibration}

Table~\ref{tab:speed_memory_calibration_SOTA_comparison_appendix} shows that GRINQH's calibration latency is significantly lower than data-dependent quantization methods like GPTQ and AWQ. 
For the Llama-3-8B-Instruct model, GRINQH achieves a $\approx 15.7\times$ speedup over AWQ and $\approx8.5\times$ over GPTQ. 

Importantly, GRINQH is quantization-agnostic; it calibrates optimal thresholds for a target precision distribution on the quantized model, while the prior quantization step is independent of the threshold calibration itself. 
In our experiments, we mostly use Round-to-Nearest (RTN) for the quantization, which adds negligible overhead to the total setup time. 
Furthermore, since the perplexity on the calibration set is obtained as a byproduct of the calibration run, it can be utilized for zero-cost hyperparameter selection (see Section~\ref{app:HyperparameterGuide}). 
This allows for 10--15 iterative optimization runs while remaining competitive with the time required for a single AWQ calibration.

Finally, we note that our current calibration pipeline is entirely unoptimized, leaving significant room for further speedups through, e.g., sample-level parallelization.
\begin{threeparttable}
    \centering
    \caption{Preparation speed and memory benchmarks for Llama-3.1-8B on a single NVIDIA RTX 6000 Blackwell. RTN, GPTQ, and AWQ are implemented in \texttt{llm-compressor}; GRINQH benchmarks only the threshold calibration phase within \texttt{vLLM}, where a single precision distribution (single datapoint) determines the thresholds. GRINQH enables near-instantaneous model preparation with significant speedups over SOTA baselines.}

    \begin{tabular*}{\textwidth}{l @{\extracolsep{\fill}} r r r} 
    \toprule
    Method & Prep Time (s) & Peak VRAM (GiB) & Relative Time \\
    \midrule
    RTN    & 4.19 ± 0.08   & 16.44 ± 0.08    & 1.00x \\
    GPTQ   & 363.42 ± 1.38 & 16.37 ± 0.00    & 86.83x \\
    AWQ    & 673.14 ± 5.98 & 21.82 ± 0.00    & 160.83x \\
    GRINQH (single datapoint, no quant) & 42.82 ± 0.39  & 40.94 ± 0.00    & 10.23x \\
    \bottomrule
    \end{tabular*} \label{tab:speed_memory_calibration_SOTA_comparison_appendix}
\end{threeparttable}

Table~\ref{tab:speed_memory_calibration_all_models_appendix} demonstrates the scalability of GRINQH across the Llama3 and Qwen3 families. 
Both preparation time and peak VRAM scale sub-linearly with model size, confirming that the method remains viable for large-scale calibration without prohibitive hardware requirements.

\begin{table}[h]
\centering
\caption{Preparation speed and memory benchmarks for Llama3 and Qwen3 model families on a single NVIDIA RTX 6000 Blackwell. Benchmarks cover the threshold calibration phase within \texttt{vLLM}. Both preparation time and peak VRAM scale sub-linearly with model size, demonstrating the scalability of the calibration method.}
\resizebox{\textwidth}{!}{
    \begin{tabular}{lccc|cccc}
    \toprule
    Model & \multicolumn{3}{c}{Llama3} & \multicolumn{4}{c}{Qwen3} \\
    Size & 1B & 3B & 8B & 0.6B & 1.7B & 4B & 8B \\
    \midrule
    Prep Time (s) & 6.64 ± 0.02 & 17.22 ± 0.26 & 42.82 ± 0.39 & 4.08 ± 0.12 & 9.22 ± 0.12 & 23.48 ± 0.25 & 42.14 ± 0.77 \\
    Relative Time & 1x & 2.59x & 6.45x & 1x & 2.26x & 5.75x & 10.32x \\
    Peak VRAM (GiB) & 13.90 ± 0.00 & 27.93 ± 0.00 & 40.94 ± 0.00 & 12.81 ± 0.00 & 14.63 ± 0.00 & 20.85 ± 0.00 & 31.82 ± 0.00 \\
    Relative Peak VRAM & 1x & 2.01x & 2.95x & 1x & 1.14x & 1.63x & 2.48x \\
    \bottomrule
    \end{tabular} 
}
\label{tab:speed_memory_calibration_all_models_appendix}
\end{table}

\subsection{Representativeness of Calibration Set} \label{app:RepresentativenessCalibration}

\begin{figure}[h]
    \centering
    \includegraphics[width=\textwidth]{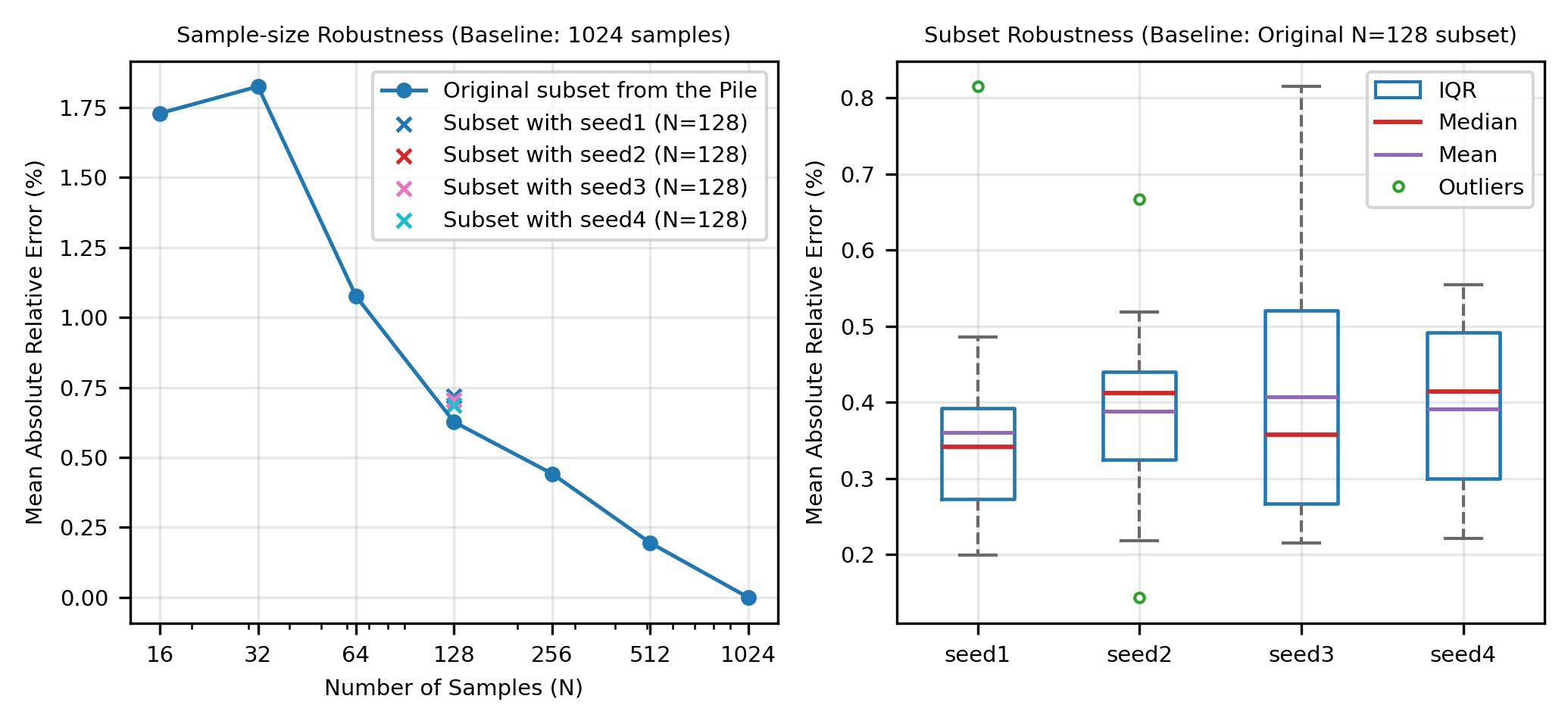}
    \caption{Robustness of calibrated thresholds to sample size and subset variation for Llama-3.1-8B-Instruct. \textit{Left (A):} MARE of thresholds relative to a 1024-sample baseline. Crosses denote per-seed variation for $N=128$. \textit{Right (B):} MARE across random subsets relative to the reference. The method remains robust across both sample size and subset selection.}
    \label{fig:calibration_robustness_appendix}
\end{figure}

We evaluate the robustness of the calibration process across varying sample sizes and subsets of the Pile (uncopyrighted) dataset. 
By sweeping calibration set sizes from 16 to 1024 samples, we observe that at a size of 128 samples (our chosen setup for baseline parity), the mean absolute relative error (MARE) of the calibrated thresholds compared to the 1024-sample reference drops below $1\%$ (Fig.~\ref{fig:calibration_robustness_appendix}A).

To assess subset sensitivity, we repeated the calibration across five random sets of 128 samples. 
The results demonstrate minimal variance, with threshold MARE consistently remaining below 1\% (see crosses in Fig.~\ref{fig:calibration_robustness_appendix}A). 
The negligible variation across subsets (Fig.~\ref{fig:calibration_robustness_appendix}B) confirms that the method is not overly sensitive to the specific choice of calibration data.

\subsection{Calibration Fidelity.} \label{app:calibrationFidelity}

We evaluate calibration fidelity by comparing target bit widths against empirically measured effective bit widths across downstream tasks. Fig.~\ref{fig:kernel_aux_data_bench}B illustrates that the measured bit widths across all models and tasks evaluated for Fig.~\ref{fig:pareto_joint} and Fig.~\ref{fig:pareto_joint_appendix} (plus common-sense reasoning benchmarks) for $b_\text{max} \in \{6, 8\}$ totaling 3632 benchmark runs closely track the ideal identity mapping, demonstrating that thresholds optimized on the calibration set generalize effectively to real-world generation. This tight alignment across diverse models and target precisions confirms that our distribution-aware mechanism successfully regulates memory traffic as intended without the need for task-specific tuning or training.

\section{Appendix: Weight Representation and Approximation Analysis} \label{app:weight_representation}

This section examines the technical design choices governing GRINQH’s weight representation and its approximation logic. Specifically, we provide a detailed comparison between hardware-efficient bit-truncation (slicing) and standard rounding (Section~\ref{app:truncationVsRounding}), demonstrating how our framework compensates for approximation shifts. Furthermore, we analyze the compatibility and performance trade-offs of integrating various nested weight formats into the GRINQH pipeline (Section~\ref{app:evaluationOfNestedWeightFormats}).

\subsection{Truncation vs. Rounding}\label{app:truncationVsRounding}

While standard uniform quantization typically employs rounding to minimize approximation error, GRINQH utilizes bit-truncation (slicing). True rounding would require loading an additional bit of precision to determine the rounding direction, negating the memory traffic savings achieved by reducing bit widths. Instead, GRINQH employs truncation and compensates for the resulting mean shift by applying a pre-computed truncation bias (see Section~\ref{sec:hierarchicalBitSlicing} for more details).

\begin{figure}[h]
    \centering
    \includegraphics[width=\textwidth]{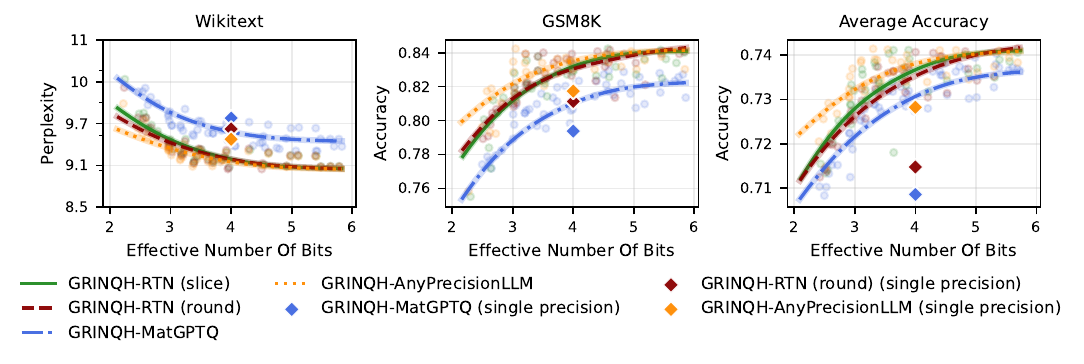}
    \caption{Quantization method comparison across key metrics on Llama3 8B. Performance metrics (perplexity on wikitext, exact match accuracy on GSM8K, and average task accuracy) versus effective number of bits for four quantization approaches: GRINQH-RTN (slice), GRINQH-RTN (round), GRINQH-MatGPTQ, and GRINQH-AnyPrecisionLLM. Curves represent smoothed trends via Gaussian-kernel estimation over multiple precision distribution configurations. Single precision distributions are shown as diamonds; mixed-precision distributions are as circles.}
    \label{fig:rounding_vs_slicing_appendix}
\end{figure}

As shown in Tab.~\ref{tab:slice_vs_round_all_models_appendix}, substituting rounding with truncation in the GRINQH framework results in negligible variations in average accuracy. This trend is further illustrated across varying effective bit widths in Fig.~\ref{fig:rounding_vs_slicing_appendix}, where truncation and rounding curves largely overlap. While rounding remains slightly superior in the extreme 2-bit regime, the two methods achieve nearly identical average performance.

Notably, this stability is unique to the mixed-precision setting. In fixed-precision configurations, truncation significantly degrades performance compared to rounding, to the point where data points fall outside the viable scale of our plots. This suggests that GRINQH’s input-adaptive precision allocation effectively mitigates the local errors introduced by truncation, allowing for hardware-friendly memory loads without sacrificing model fidelity.

\resizebox{\textwidth}{!}{
\begin{threeparttable}
\caption{Performance difference between slicing from 8-bit versus rounding weights to lower precision ($\Delta$ = slice $-$ round), averaged over 40 GRINQH configurations (different precision distributions) covering the 2.1--5.7 effective-bit range, across the Llama3 and Qwen3 families with RTN as the 8-bit baseline. Slicing from 8-bit achieves comparable performance to rounding, with differences remaining 
negligible on average across all tested models.}
\begin{tabular}{lrrrrr}
\toprule
& \multicolumn{5}{c}{$\Delta$ (= slice $-$ round)} \\
\cmidrule(lr){2-6}
Model 
& \makecell{Wikitext \\ PPL $\downarrow$} 
& \makecell{Lambada \\ PPL $\downarrow$} 
& \makecell{GSM8K \\ Acc. (\%) $\uparrow$} 
& \makecell{Average \\ Acc. (\%) $\uparrow$} 
& \makecell{Eff. \\ Bits} \\
\midrule
Llama-3.2-1B-Instruct-RTN (Max 8-bit) & 0.0196 & 0.6904 & +0.24 & -0.06 & 0.0028 \\
Llama-3.2-3B-Instruct-RTN (Max 8-bit) & 0.1392 & 0.0060 & -0.36 & -0.32 & 0.0070 \\
Llama-3.1-8B-Instruct-RTN (Max 8-bit) & 0.0531 & -0.0028 & -0.20 & +0.08 & 0.0104 \\
Qwen3-0.6B-RTN (Max 8-bit)            & 1.2720 & 5.9783 & -1.15 & +0.12 & 0.0048 \\
Qwen3-1.7B-RTN (Max 8-bit)            & 1.1128 & 0.6365 & -0.99 & -0.58 & 0.0223 \\
Qwen3-4B-RTN (Max 8-bit)              & 0.1184 & -0.0879 & -1.00 & -0.03 & 0.0074 \\
Qwen3-8B-RTN (Max 8-bit)              & 0.1313 & 0.0429 & +0.32 & -0.07 & 0.0117 \\
\bottomrule
\end{tabular} \label{tab:slice_vs_round_all_models_appendix}
\end{threeparttable}
}

\subsection{Evaluation of Nested Weight Formats} \label{app:evaluationOfNestedWeightFormats}

Beyond uniform quantization, GRINQH is compatible with nested-weight representations such as any-precision LLM \cite{AnyPrecisionLLM24}. While MatGPTQ \cite{MatGPTQ26} also employs nested weights, its loading logic requires fetching an additional bit beyond the target bit width, making it incompatible with GRINQH’s objective of minimized memory traffic.

Fig.~\ref{fig:rounding_vs_slicing_appendix} shows that MatGPTQ, which utilizes a GPTQ-based nested structure, exhibits the lowest performance in our evaluation. This highlights a fundamental representation trade-off: nested formats often sacrifice high-bit fidelity to improve low-bit quantization performance. This characteristic undermines GRINQH’s outlier protection strategy, which relies on allocating maximum precision to weights associated with high-magnitude activations to minimize or eliminate quantization error in critical channels. If the underlying nested format introduces error even at its highest precision level, the benefit of dynamic allocation is diminished.

In contrast, any-precision LLM demonstrates the potential of nested structures by outperforming baselines in the sub-4-bit regime while matching the performance of GRINQH-RTN (with slicing) at higher effective bit widths. This suggests that while nested weights can enhance efficiency, their effectiveness within a dynamic framework depends heavily on their ability to maintain high-fidelity representation for salient weights.


\end{document}